\documentclass{article}

    \usepackage[final]{neurips_2020}

\usepackage[utf8]{inputenc} 
\usepackage[T1]{fontenc}    
\usepackage{hyperref}       
\usepackage{url}            
\usepackage{booktabs}       
\usepackage{amsfonts}       
\usepackage{nicefrac}       
\usepackage{microtype}      

\usepackage{defs}

\usepackage[letterpaper]{geometry}
\usepackage[parfill]{parskip}
\PassOptionsToPackage{numbers, compress}{natbib}
\usepackage[numbers, compress]{natbib}
\usepackage{xr-hyper,refcount}

\usepackage{graphicx}
\usepackage[utf8]{inputenc} 
\usepackage[T1]{fontenc}    
\usepackage{url}            
\usepackage{booktabs}       
\usepackage{amsfonts}       
\usepackage{nicefrac}       
\usepackage{microtype}      
\usepackage[parfill]{parskip}
\usepackage{algorithm,algorithmic}
\usepackage{amsmath,amsthm,amssymb,bbm}
\usepackage{mathtools}
\usepackage{cases}
\usepackage{comment}
\usepackage{subcaption}
\usepackage{algorithm,algorithmic}
\usepackage{color}
\usepackage{appendix}
\usepackage{xspace}
\usepackage{enumitem}
\usepackage[english]{babel}

\usepackage{cleveref}
\crefformat{equation}{(#2#1#3)}
\crefrangeformat{equation}{(#3#1#4) to~(#5#2#6)}
\crefname{equation}{}{}
\Crefname{equation}{}{}

\crefname{definition}{\textbf{definition}}{definitions}
\Crefname{definition}{Definition}{Definitions}
\crefname{assumption}{\textbf{assumption}}{assumptions}
\Crefname{assumption}{Assumption}{Assumptions}
\definecolor{maroon}{RGB}{192,80,77}

\newtheorem{theorem}{Theorem}
\newtheorem{lemma}[theorem]{Lemma}

\newtheorem{corollary}[theorem]{Corollary}
\newtheorem{definition}[theorem]{Definition}

\usepackage{thmtools, thm-restate}

\title{Sample Complexity of Uniform Convergence\\ for Multicalibration}
\author{
    Eliran Shabat\thanks{
First two authors have equal contribution.
} \\
    Tel Aviv University\\
    \texttt{shabat.eliran@gmail.com} \\
    \And
    Lee Cohen$^*$ \\
    Tel Aviv University \\
    \texttt{leecohencs@gmail.com} \\
  \And
    Yishay Mansour \\
    Tel Aviv University and\\
      Google Research \\
    \texttt{mansour.yishay@gmail.com} \\
}

\begin{document}

\maketitle

\begin{abstract}
There is a growing interest in societal concerns in machine learning systems, %
especially in fairness.
Multicalibration gives a comprehensive methodology to address group fairness.
In this work, we address the multicalibration error  and decouple it from the prediction error.
The importance of decoupling the fairness metric (multicalibration) and the accuracy (prediction error) is due to the inherent trade-off between the two, and the societal decision  regarding %
the ``right tradeoff'' (as imposed many times by regulators). %
Our work gives sample complexity bounds for uniform convergence guarantees of multicalibration error, which implies that regardless of the accuracy, we can guarantee that the empirical and (true) 
multicalibration errors are close.
We emphasize that our results: (1) are more general than previous bounds, as they apply to both agnostic and realizable settings, and do not rely on a specific type of algorithm (such as differentially private), %
(2) improve over previous multicalibration sample complexity bounds and (3) implies uniform convergence guarantees for the classical calibration error.

\end{abstract}
\section{Introduction}
\label{sec:intro}
%
Data driven algorithms influence our everyday lives. While they introduce significant achievements in face recognition, to recommender systems and machine translation, they come at a price.
When deployed for predicting outcomes that concern individuals, such as repaying a loan, surviving surgery, or skipping bail, predictive systems are prone to accuracy disparities between different social groups that often induce discriminatory results.
These significant societal issues arise due to a variety of reasons:
problematic analysis, unrepresentative data 
and even inherited biases against certain social groups due to historical prejudices.
At a high level, there are two separate notions of fairness: \textit{individual fairness} and \textit{group fairness}. Individual fairness is aimed to guarantee fair prediction to each given individual, while group fairness aggregates statistics of certain subpopulations, and compares them. There is a variety of fairness notions for group fairness, such as demographic parity, equalized odds, equalized opportunity, and more (see \cite{FairnessBook2019}). Our main focus would be on multicalibration criteria for group fairness \cite{Hebert-Johnson2018}.
Multicalibration of a predictor is defined as follows.
There is a prespecified set of subpopulations of interest.
The predictor returns a value for each individual (which can be interpreted as a probability). The multicalibration requires that for any ``large'' subpopulation, and for any value which is predicted ``frequently'' on that subpopulation, the predicted value and average realized values would be close on this subpopulation.
Note that calibration addresses the relationship between the predicted and average realized values, and is generally unrelated to the prediction quality. For example, if a population is half positive and half negative, a predictor that predicts for every individual a value of $0.5$ is perfectly calibrated but has poor accuracy.
The work of \cite{Hebert-Johnson2018} proposes a specific algorithm to find a multicalibrated predictor and derived its sample complexity. The work of  \cite{Liu2018} related the calibration error to the prediction loss, specifically, it bounds the calibration error as a function of the difference between the predictor loss and the Bayes optimal prediction loss. Their bound implies that in a realizable setting, where the Bayes optimal hypothesis is in the class, using ERM
 yields a vanishing calibration error, but in an agnostic setting this does not hold.
With the motivation of fairness in mind, it is important to differentiate between the prediction loss and the calibration error. In many situations, the society (through regulators) might sacrifice prediction loss to improve fairness, and the right trade-off between them may be task dependent. On the other hand, calibration imposes self-consistency, namely, that predicted values and the average realized values should be similar for any protected group. In particular, there is no reason to prefer un-calibrated predictors over calibrated ones, assuming they have the same prediction loss.
An important concept in this regard is uniform convergence. We would like to guarantee that the multicalibration error on the sample and the true multicalibration error are similar.
This will allow society to rule-out un-calibrated predictors when optimizing over accuracy and other objectives that might depend on the context and the regulator.\\
Our main results in this work are sample bounds that guarantee uniform convergence of a given class of predictors. We start by deriving a sample bound for the case of a finite hypothesis class, and derive a sample complexity bound which is logarithmic in the size of the hypothesis class.
Later, for an infinite hypothesis class, we derive a sample bound that depends on the \textit{graph dimension} of the class (which is an extension of the VC dimension for multiclass predictions). Finally, we derive a lower bound on the sample size required.\\
Technically, an important challenge in deriving the uniform convergence bounds is that the multicalibration error depends, not only on the correct labeling but also on the predictions by the hypothesis, similar in spirit to the internal regret notion in online learning. We remark that these techniques are suitable to reproduce generalization bounds for other complex measures such as F-score.

We stress that in contrast to previous works that either attained specific efficient algorithms for finding calibrated predictors \cite{Hebert-Johnson2018} or provided tight connections between calibration error and prediction loss (mainly in the realizable case) \cite{Liu2019}, we take a different approach. We concentrate on the statistical aspects of generalization bounds rather than algorithmic ones, and similar to much of the generalization literature in machine learning derive generalization bounds over calibration error for \textit{any} predictor class with a finite size or a finite graph dimension. 

Nevertheless, our work does have algorithmic implications. 
For example, similarly to running ERM, running empirical multicalibration risk minimization over a hypothesis class with bounded complexity $\HH$ and ``large enough'' training set, would output a nearly-multicalibrated predictor, assuming one exists.
%
We guarantee that the empirical and true errors of this predictor would be similar, and derive the required sample size either as a function of the logarithm of the size of the predictor class or of its finite graph dimension.
%
Our bounds improve over previous sample complexity bounds and also apply in more general settings (e.g., agnostic learning). So while multicalibration uniform convergence is not formally necessary for learning multicalibrated predictors, the advantage of our approach is that the learner remains with the freedom to choose any optimization objectives or algorithms, and would still get a good estimation of the calibration error.
To the best of our knowledge, this also introduces the first uniform convergence results w.r.t. calibration as a general notion (i.e., even not as a fairness notion).

\noindent{\bf Related work:}
Calibration has been extensively studied in machine learning, statistics and economics \cite{Foster1998,
BlumMansour2007,FosterHart2018}, and as a notion of fairness dates back to the 1960s \cite{Cleary1968}.
More recently, the machine learning community adapted calibration as an anti-discrimination tool and studied it and the relationship between it and other fairness criteria \cite{Chouldechova2017, Corbett-Davies2017, Kleinberg2017, Pleiss2017,Liu2017}.
There is a variety of fairness criteria, other than calibration, which address societal concerns that arise in machine learning.
Fairness notions have two major categories. \textit{Individual-fairness}, that are based on similarity metric between individuals and require that similar individuals will be treated similarly \cite{Dwork2012}. \textit{Group-fairness}, such as demographic-parity and equalized-odds, are defined with respect to statistics of subpopulations \cite{FairnessBook2019}.
Generalization and uniform convergence are well-explored topics in machine learning, and usually assume some sort of hypotheses class complexity measures, such as VC-dimension, Rademacher complexity, Graph-dimension and Natarajan-dimension \cite{BenDavid1995, Daniely2011, mlBook}. In this work we build on these classic measures to derive our bounds.
Generalization of fairness criteria is a topic that receives great attention recently.
The works of \cite{Kim2018, YonaRothblum2018} define metric notions that are based on \cite{Dwork2012} and derive generalization guarantees. 
Other works relax the assumption of a known fairness metric and derive generalization with respect to Individual Fairness based on oracle queries that simulate human judgments \cite{Gillen2018, bechavod2020metricfree, ilvento2020}.
Bounds for alternative fairness notions, such as equalized-odds, gerrymandering, multi-accuracy, and envy-free appear in \cite{Woodworth2017, Kearns2018, Kim2019, Balcan2019}.
We remark that this work does not provide generalization bounds for margin classifiers in the context of fairness, and we leave it for future work.

Multicalibration is a group-fairness notion that requires calibration to hold simultaneously on multiple  subpopulations \cite{Hebert-Johnson2018}.
They proposed a polynomial-time differentially-private algorithm that learns a multicalibrated predictor from samples in agnostic setup.
A byproduct of their choice of Differently Private algorithm is that their algorithm and analysis is limited to a finite domain.
Our work provides generalization uniform convergence bounds that are independent of the algorithm that generates them, and also improve their sample bounds.
%
%
The work of \cite{Liu2019} bounds the calibration error by the square-root of the gap between its expected loss and the Bayes-optimal loss, for a broad class of loss functions.
While in realizable settings this gap is vanishing, in  agnostic settings this gap can be substantial.
Our results do not depend on the hypothesis' loss to bound the calibration error, which allows us to give guarantees in the agnostic settings as well.

\section{Model and Preliminaries}
\label{mo}

Let $\X$ be any finite or countable domain (i.e., $\X$ is a population and each domain point encodes an individual) and let $\{0,1\}$ be the set of possible \textit{outcomes}. Let $D$ be a probability distribution over $\X \times \{0,1\}$, i.e., a joint distribution over domain points and their outcomes.
Intuitively, given pairs $(x_i,y_i)$, we assume that outcomes $y_i\in \{0,1\}$ are the  realizations of underlying random sampling from independent Bernoulli distributions with (unknown) parameters $p^*(x_i)\in [0,1]$. 
The goal of the learner is to predict the (unknown) parameters $p^*(x_i)$, given a domain point $x_i$.
Let $\mathcal{Y} \subseteq [0,1]$ be the set of possible \textit{predictions values}. 
A \textit{predictor} (hypothesis) $h$ is a function that maps domain points from $\X$ to prediction values $v\in \Y$.
A set of  predictors $h: \X \rightarrow \Y$ is a \textit{predictor class} and denoted by $\HH$.
Let $\Gamma = \{U_1, ..., U_{|\Gamma|}\}$ be a finite collection of subpopulations (possibly overlapping) from the domain $\mathcal{X}$ (technically, $\Gamma$ is a collection of subsets of $\X$). 
Throughout this paper, we will distinguish between the case where $\mathcal{Y}$ is a finite subset of $[0,1]$ and the case where $\Y=[0,1]$ (continuous).
Both cases depart from the classical binary settings where $\Y=\{0,1\}$, as predictors can return any prediction value $v\in \Y$ (e.g., $v=0.3$). 
We define $\Lambda$ to be a \textit{partition} of $\Y$ into a finite number of subsets, that would have different representations in the continuous and finite cases.
For the continuous case where $\Y=[0,1]$, we would partition $\Y$ into a finite set of intervals using a \textit{partition parameter} $\lambda\in (0,1]$ that would determinate the lengths of the intervals. Namely,  $\Lambda_\lambda:= \{\{I_{j}\}_{j=0}^{\frac{1}{\lambda}-1}\}$, where $I_{j}=
[j\lambda,(j+1)\lambda)$.
When $\Y$ is finite, $\Lambda$ would be a set of singletons: $\Lambda=\{\{v\}: v\in \Y\}$ and $h(x)\in I=\{v\}$ is equivalent to $h(x)=v$.
%
\begin{definition}[Calibration error]
The {\em calibration error} of predictor $h\in \HH$  w.r.t. a subpopulation $U\in \Gamma$ and an interval $I\subseteq [0,1]$,  denoted by $c(h,U,I)$ 
is the difference between the expectations of $y$ and $h(x)$, conditioned on domain points from $U$ that $h$ maps to values in $I$.  
I.e., 
\begin{align*}
    c(h,U,I) &:= \Ex_{D}\condsb{y}{x \in U, h(x)\in I} 
              - \Ex_{D}\condsb{h(x)}{x \in U,  h(x)\in I}
\end{align*}
\end{definition}
Notice that for the case where $\Y$ is finite, we can rewrite the expected calibration error as
\begin{align*}
    c(h,U,I = \{v\}) &= \Ex_{D}\condsb{y}{x \in U, h(x)= v} - v
\end{align*}
%
Since calibration error of predictors is a measure with respect to a specific pair of subpopulation $U$ and an interval $I$, we would like to have a notion that captures  ``well-calibrated'' predictors on ``large enough'' subpopulations and ``significant enough'' intervals $I$ that $h$ maps domain points (individuals) to, as formalized in the following definition.
\begin{definition}[Category]
A {\em category} is a pair $(U, I)$ of a subpopulation $U \in \Gamma$ and an interval $I \in \Lambda$.
We say that a category $(U, I)$ is {\em interesting} according to predictor $h$ and parameters $\gamma, \psi \in (0,1]$, if $\Prob_D[x \in U] \geq \gamma$ and $\Prob_D\condsb{h(x) \in I}{x \in U} \geq \psi$.
\end{definition}
We focus on predictors with calibration error of at most $\alpha$ for any interesting category. 

\begin{definition}[$(\alpha, \gamma, \psi)$--multicalibrated predictor]
A predictor $h\in \HH$ is $(\alpha, \gamma, \psi)$--{\em multicalibrated},
if for every
interesting category $(U,I)$ according to $h$, $\gamma$ and $\psi$,
the absolute value of the calibration error of $h$ w.r.t. the category $(U,I)$ is at most $\alpha$, i.e., $\bigl\lvert c(h,U,I)  \bigr\rvert \leq \alpha$.
\end{definition}
We define empirical versions for calibration error and $(\alpha, \gamma, \psi)$--multicalibrated predictor.
%
\begin{definition}[Empirical Calibration error]
Let $(U,I)$ be a category and let $S^m = \{(x_1, y_1), ..., (x_m, y_m)\}$ be a training set of $m$ samples drawn i.i.d. from $D$. The {\em empirical calibration error} of a predictor $h\in \HH$ w.r.t. $(U,I)$ and $S$ is:
\[
    \hat{c}(h,U,I,S) := \sum_{i=1}^{m}\frac{\indicator{x_i \in U, h(x_i) \in I} }{\sum_{j=1}^{m}{\indicator{x_j \in U, h(x_j) \in I}}}y_i 
    - \sum_{i=1}^{m}\frac{\indicator{x_i \in U, h(x_i) \in I}} {\sum_{j=1}^{m}{\indicator{x_j \in U, h(x_j)\in I}}}h(x_i), 
\]
where $\indicator{\cdot}$ is the indicator function.
\end{definition}

Notice that when $\mathcal{Y}$ is finite, since $h(x)\in \{v\}$ is equivalent to $h(x)=v$, we can re-write the empirical calibration error as:
$\hat{c}(h,U,I = \{v\},S) := 
    \sum_{i=1}^{m}\frac{\indicator{x_i \in U, h(x_i) = v} }{\sum_{j=1}^{m}{\indicator{x_j \in U, h(x_j) = v}}}y_i - v$.
%
\begin{definition}[$(\alpha, \gamma, \psi)$--Empirically multicalibrated predictor]
A predictor $h\in \HH$ is $(\alpha, \gamma, \psi)$--{\em empirically multicalibrated} on a sample $S$ of i.i.d examples from $D$,
if for every interesting category $(U,I)$ according to $h$, $\gamma$ and $\psi$, we have
$\bigl\lvert \hat{c}(h,U,I,S)  \bigr\rvert \leq \alpha
$.
\end{definition}

We assume that the predictors are taken from some predictor class $\mathcal{H}$.
Our main goal is to derive sample bounds for the empirical calibration error to ``generalize well'' for every $h \in \mathcal{H}$ and every interesting category.
We formalize it as follows.


\begin{definition}[Multicalibration Uniform Convergence] 
A predictor class $\mathcal{H} \subseteq \mathcal{Y}^\mathcal{X}$ has the {\em multicalibration uniform convergence} property (w.r.t. collection $\Gamma$) if there exist a function $m_{\HH}^{mc}(\epsilon,\delta,\gamma, \psi) \in \mathbb{N}$, for $\epsilon,\delta,\gamma, \psi \in (0,1]$, such that for every distribution $D$ over $\mathcal{X} \times \{0,1\}$, if $S^m = \left\{(x_1, y_1), \cdots, (x_m, y_m)\right\}$ is a training set of $m \geq m_{\HH}^{mc}(\epsilon,\delta,\gamma, \psi)$ examples drawn i.i.d. from $D$, 
then for every $h \in \HH$ and every interesting category $(U,I)$ according to $h$, $\gamma$ and $\psi$,
the difference between the calibration error and the empirical calibration error is at most $\epsilon$ with probability of at least $1-\delta$, i.e.,  
$\Pr_D[|\hat{c}(h,U,I,S^m) - c(h,U,I)| \leq \epsilon] > 1 - \delta$.
\end{definition}

We emphasize that the property of multicalibration uniform convergence w.r.t. a predictor class $\mathcal{H}$ is neither a necessary nor sufficient for having multicalibrated predictors $h\in \mathcal{H}$.
Namely, having uniform convergence property implies only that the empirical and true errors are similar, but does not imply that they are small. In addition, having a predictor with zero multicalibration error (realizability) does not imply anything about the generalization multicalibration error. For example, if $\mathcal{H}$ contains all the possible predictors, there will clearly be a zero empirical error predictor who's true multicalibration error is very high.\\
When $\mathcal{H}$ is an infinite predictor class, we can achieve generalization by assuming a finite complexity measure. VC-dimension (the definition appears in section \ref{app:usefulThms} in the Supplementary Material) 
measures the complexity of binary hypothesis classes. In this work, we rephrase the generalization problem of multicalibration in terms of multiple generalization problems of binary hypothesis classes with finite VC-dimension,
and derive sample complexity bounds for it.
So our goal is to approximate the (true) calibration error by estimating it on a large sample. Namely, we would like  have a property which indicates that a large-enough sample will result a good approximation of the calibration-error for any hypothesis $h \in \mathcal{H}$ and any interesting category $(U, I)$ according to $h$. Our technique for achieving this property uses known results about binary classification. We mention the definitions of ``risk function'', ``empirical-risk function'' and ``uniform convergence for statistical learning'' (the latter appears in section \ref{app:usefulThms} in the Supplementary Material). For this purpose, $h : \mathcal{X} \rightarrow \{0,1\}$ would denote a binary hypothesis, $\ell : \mathcal{Y} \times \{0,1\} \rightarrow \mathbb{R}_+$, denotes a loss function and $D$ stays a distribution over $\mathcal{X} \times \{0,1\}$. 

\begin{definition}[Risk function, Empirical risk]
The {\em risk function}, denoted by $L_D$, is the expected loss of a hypothesis $h$ w.r.t $D$, i.e., 
$L_D(h) := \Ex_{(x,y) \sim D}[\ell(h(x), y)]$.
%
Given a random sample $S=\left((x_i, y_i)\right)_{i=1}^m$  of $m$ examples drawn i.i.d. from $D$, the {\em empirical risk} is the  average loss of $h$ over the sample $S$ i.e., $L_S(h) := \frac{1}{m}\sum_{i=1}^{m}{\ell(h(x_i),y_i)}$.
\end{definition}



Note that the definitions of uniform convergence for statistical learning and the multicalibration uniform convergence are distinct.
A major difference is that while the notion of uniform convergence for statistical learning imposes a requirement on the risk, which is defined using an expectation over a fixed underlying distribution $D$, the notion of multicalibration uniform convergence imposes a requirement on the calibration error, in which the expectation is over a conditional distribution that depends on the predictor.
When the prediction range, $\mathcal{Y}$, is discrete, 
we consider the standard multiclass complexity notion  graph-dimension, which is define as follows.

\begin{definition}[Graph Dimension]
Let $\mathcal{H} \subseteq \mathcal{Y}^\mathcal{X}$ be a hypothesis class from domain $\mathcal{X}$ to a finite set $\mathcal{Y}$ and let $S \subseteq \mathcal{X}$. We say that $\mathcal{H}$ G-shatters $S$ if there exists a function $f : S \rightarrow \mathcal{Y}$ such that for every $T \subseteq S$ there exists a hypothesis $h \in \mathcal{H}$ such that
$\forall x \in S: h(x) = f(x) \iff x \in T$.
The graph dimension of $\mathcal{H}$, denoted $d_G(\mathcal{H})$, is the maximal cardinality of a set that is G-shattered by $\mathcal{H}$.
\end{definition}

\section{Our Contributions}
We derive two upper bounds. The first is for a finite predictor class, in which we discretize $\Y=[0,1]$ into $\Lambda_\lambda$ and derive a bound which depends logarithmicly on $\lambda^{-1}$. 
We complement our upper bounds with the following lower bound result.
\begin{restatable}{theorem}{finiteThm}
\label{main:thm:finite_gen_thm}%
Let $\mathcal{H} \subseteq \mathcal{Y}^\mathcal{X}$ be a finite predictor class. 
Then, $\mathcal{H}$ has the uniform multicalibration convergence property with
$m_{\mathcal{H}}^{mc}(\epsilon, \delta, \gamma, \psi)
=O\left(
\frac{1}{ \epsilon^2 \gamma \psi}
\log \left(|\Gamma||\mathcal{H}| /\delta \lambda \right) \right)$.
\end{restatable}
%
\begin{restatable}{theorem}{graphThm}
\label{main:thm:graph_ub}
Let $\mathcal{H} \subseteq \mathcal{Y}^\mathcal{X}$ be an infinite predictor class from domain $\mathcal{X}$ to a discrete prediction set $\mathcal{Y}$ with finite graph-dimension $d_G(\mathcal{H}) \leq d$, then $\HH$
has the uniform multicalibration convergence property with 
$m_{\mathcal{H}}^{mc}(\epsilon, \delta, \gamma, \psi) =O\left(
 \frac{1}{\epsilon^2 \psi^2 \gamma}\left(d + \log\left( |\Gamma||\mathcal{Y}|/\delta \right)\right) \right)$.
\end{restatable}
\begin{restatable}{theorem}{lowerBoundThm}
\label{main:thm:lowerBound}
Let $\HH$ be a finite predictor class or an infinite predictor class with finite graph-dimension $d_G(\mathcal{H}) \leq d$. Then, $\HH$ has  multicalibration uniform convergence with $m(\epsilon,\delta,\psi,\gamma)= \Omega(\frac{1}{\psi\gamma\epsilon^2}\ln (1/\delta))$ samples. 
\end{restatable}


Rewriting the sample bound of \cite{Hebert-Johnson2018} using our parameters, they have
${O}\left(\frac{1}{\epsilon^3\cdot\psi^{3/2}\cdot\gamma^{3/2}}\log(\frac{|\Gamma|}{\epsilon\cdot\gamma\cdot\delta})\right)$.
Comparing the bounds, the most important difference is the dependency on $\epsilon$, the generalization error. They have a dependency of $\epsilon^{-3}$, while we have of $\epsilon^{-2}$, which is tight due to our lower bound. 
For the dependency on $\gamma$, they have $\gamma^{-3/2}$, while we have $\gamma^{-1}$, which is also tight. 
For the dependency on $\psi$, they have $\psi^{-3/2}$, while we have $\psi^{-1}$ for a finite hypothesis class (which is tight due to our lower bound) and $\psi^{-2}$ for an infinite hypothesis class.
Finally, recall that the bound of \cite{Hebert-Johnson2018} applies only to their algorithm and since it is a differentially private algorithm, it requires the domain $\X$ to be finite, while our results apply to continuous domains as well. 
Note that having $(\alpha, \gamma, \psi)$-- empirically multicalibrated predictor on large random sample, guarantees that, with high probability,  it is also $(\alpha + \epsilon, \gamma, \psi)$--mutlicalibrated with respect to the underlying distribution, where $\epsilon$ is the generalization error that depends on the sample size.
For brevity, we only overview our proof techniques, and provide full proofs in the Supplementary material.



\section{Finite Predictor Classes}
\label{sec:Finite}
We start by analyzing the case in which $\mathcal{H}$ is finite and the prediction set $\mathcal{Y}$ is continuous.
In this setup, we will utilize the fact that a finite $\mathcal{H}$ implies a finite number of categories, i.e., a partition of the population $\mathcal{X}$ into sub-groups according to $\mathcal{H}$, $\Gamma$ and $\Lambda_{\lambda}$. 
This fact, using the union-bound,  will allow us to translate any confidence we have over a single interesting category, to a confidence over all interesting categories while only suffering a logarithmic increase in the number of possible categories.
Recall that in this setup, the prediction-intervals set, $\Lambda$, is a partition of $\Y=[0,1]$ into a finite set of intervals of length $\lambda$, namely, $\Lambda =\{I_j\}_{j=0}^{\frac{1}{\lambda}-1}= \{[j\lambda,(j+1)\lambda)\}_{j=0}^{\frac{1}{\lambda}-1}$.





Our upper bound analysis will use the following intuition.
Assuming a large sample, with high probability, each interesting category would have a ``large enough'' sub-sample, which would yield 
a good approximation of it's calibration error with high probability.




\finiteThm*


In the proof of Theorem \ref{main:thm:finite_gen_thm} we use the relative Chernoff inequality (Lemma \ref{rel_chernoff_thm}) and union bound to guarantee,with probability of at least $1 - \delta/2$, a large sub-sample for every predictor $h \in \mathcal{H}$ and for every interesting category $(U, I)$ according to $h$.
Then, we use the absolute Chernoff inequality (Lemma \ref{abs_chernoff_thm}) to show that with probability at least $1 - \delta/2$, for every $h \in \mathcal{H}$ and every interesting category $(U, I)$ according to $h$, the empirical calibration error does not deviate from the (true) calibration error by more than $\epsilon$. 
The following corollary indicates that having $(\alpha, \gamma, \psi)$-- empirically multicalibrated predictor on a large random sample, guarantees it is also $(\alpha + \epsilon, \gamma, \psi)$--mutlicalibrated with respect to the underlying distribution with high probability, where $\epsilon$ is a generalization error that depends on the sample size.
It follows immediately from Theorem \ref{main:thm:finite_gen_thm}.
\begin{corollary}\label{cor:finite}
Let $\mathcal{H} \subseteq \mathcal{Y}^\mathcal{X}$ be a finite predictor class and let $D$ be a distribution over $\mathcal{X} \times \{0,1\}$.
Let $S$ be a random sample of $m$ examples drawn i.i.d. from $D$ and let $h \in \mathcal{H}$ be $(\alpha, \gamma, \psi)$-- empirically multicalibrated predictor on $S$.
Then, for any $\epsilon, \delta \in (0,1)$, if 
$m \geq 
\frac{ 8 }{ \epsilon^2 \gamma \psi } \log \left( \frac{ 8|\Gamma||\mathcal{H}| }{ \delta \lambda } \right)$,
then with probability at least $1 - \delta$, $h$ is $(\alpha + \epsilon, \gamma, \psi)$--multicalibrated w.r.t. the underlying distribution $D$.
\end{corollary}

\ignore{
\begin{proof}
Let $S^m = \{(x_1, y_1), ..., (x_m, y_m)\}$ be a random sample of size $m \geq m_{\mathcal{H}}(\epsilon, \delta, \psi, \gamma, \lambda)$ labeled examples drawn i.i.d. according to $D$.

For convenience, throughout the proof we use the following notations. We first define the quantities with respect to the distribution. For a given hypothesis $h\in H$,  group $U\in \Gamma$ and interval $I\in \Lambda$, we are interested in the sub-population which belongs to $U$ and for which $h$ prediction is in $I$, i.e.,  $[x\in U,h(x)\in I]$.
For this sub-population we define:
$p(h,U,I)$ the probability of being in this sub-population, $\mu_y(h,U,I)$ the average $y$ value in the sub-population, and $\mu_h(h,U,I)$, the average prediction, i.e., $h(x)$. The three measures are with respect to the true distribution $D$.
Formally,
\begin{align*}
    & p(h,U,I) := \Prob_{D}[x \in U, h(x) \in I] \\
    & \mu_y(h,U,I) := \Ex_D\condsb{y}{x \in U, h(x) \in I} \\
    & \mu_h(h,U,I) := \Ex_{D}\condsb{h(x)}{x \in U, h(x) \in I}
\end{align*}
Similarly we denote the three empirical quantities with respect to the sample. Namely, we denote by $\hat{n}(h,U,I,S)$, $\hat{\mu}_y(h,U,I,S)$ and $\hat{\mu}_h(h,U,I,S)$ the number of samples, empirical outcome and empirical prediction, of the sub-population $[x\in U,h(x)\in I]$.
Formally,
\begin{align*}
    & \hat{n}(h,U,I,S) := \sum_{i=1}^{m}{\indicator{x_i \in U, h(x_i) \in I}} \\
    & \hat{\mu}_y(h,U,I,S) := \sum_{i=1}^{m}\frac{{\indicator{x_i \in U, h(x_i) \in I} }}{\hat{n}(h,I,U,S)} y_i \\
    & \hat{\mu}_h(h,U,I,S) := \sum_{i=1}^{m} \frac{{\indicator{x_i \in U, h(x_i) \in I} }}{\hat{n}(h,I,U,S)}h(x_i)
\end{align*}

Then, the calibration error and the empirical calibration error can be expressed as:
\begin{align*}
    & c(h,U,I) = \mu_y(h,U,I) - \mu_h(h,U,I) \\
    & \hat{c}(h,U,I,S) = \hat{\mu}_y(h,U,I,S) - \hat{\mu}_h(h,U,I,S)
\end{align*}

An interesting category of a predictor $h \in \mathcal{H}$ is a tuple $(U \in \Gamma,I \in \Lambda)$, such that the probability of the group $U$ is at least $\gamma$ and the probability of the sub-population that is mapped by predictor $h$ to the interval $I$, conditioned on $U$, is at least $\psi$. We denote by $C_h$ the collection of all interesting categories of predictor $h$, namely,
\begin{align*}
    C_h := &\biggl\{(U,I) : U \in \Gamma, I \in \Lambda, \Prob_{D}[x \in U] \geq \gamma, \\
           &\quad \Prob_{D}\condsb{h(x) \in I}{x \in U} \geq \psi \biggr\}
\end{align*}

Note that every interesting category $(U,I) \in C_h$ has a probability of at least $\gamma \psi$, namely, for every $h \in \mathcal{H}$ and for any interesting category $(U,I) \in C_h$:
\begin{align*}
    &\Prob_{x \sim D}[x \in U, h(x) \in I] \\
    &\quad = \Prob_{x \sim D}\condsb{h(x) \in I}{x \in U} \cdot \Prob_{x \sim D}[x \in U] \geq \gamma \psi
\end{align*}

We define a ``bad'' event $B^m$ over the samples, as the event there exist some predictor and some interesting category for which the generalization error is larger than $\epsilon$.
\begin{align*}
    B^m := & \biggl\{ S \in (\mathcal{X} \times \{0,1\})^m : \\
    &\quad \exists h \in \mathcal{H} ,\exists (U,I) \in C_h : \\
    &\quad\quad |\hat{c}(h,U,I,S) - c(h,U,I)| > \epsilon \biggr\}
\end{align*}

Bounding the probability that $S^m \in B^m$ by $\delta$ implies the theorem.
In order to do so, we would like to have a ``large enough'' induced sample in every interesting category.
For this purpose, we define the ``good'' event, $G^{m,l}$, as the event that indicates that for every predictor, each interesting category has at least $l$ samples.
\begin{align*}
    G^{m,l} := &\biggl\{ S \in (\mathcal{X} \times \{0,1\})^m : \\
    &\quad \forall h \in \mathcal{H}, \forall (U,I \in C_h: \hat{n}(h,U,I,S) \geq l \biggr\}
\end{align*}

We will later set $l$ to achieve $\epsilon$-accurate approximation with confidence $\delta$ later. 
Note that $G^{m,l}$ is not the complement of $B^m$.

According to the law of total probability the following holds:
\begin{align*}
    \Prob[B^m] &= \Prob\condsb{B^m}{G^{m,l}}\Prob\left[G^{m,l}\right] \\
    &\quad + \Prob\condsb{B^m}{\overline{G^{m,l}}} \Prob\left[\overline{G^{m,l}}\right] \\
    &\leq  \Prob\condsb{B^m}{G^{m,l}} + \Prob\left[\overline{G^{m,l}}\right]
\end{align*}
We would like to bound each of the probabilities\\
$\condprob{B^m}{G^{m,l}}$ and $\Prob[\overline{G^{m,l}}]$ by $\delta/2$, in order to bound the probability of $B^m$ by $\delta$.

We start by bounding $\Prob\condsb{S^m \in B^m}{S^m \in G^{m,l}}$.
By using the union bound:
\begin{align*}
    &\Prob\condsb{S^m \in B^m}{S^m \in G^{m,l}} \\
    &= \Prob\biggl[ \exists h \in \mathcal{H} ,\exists (U,I) \in C_h : \\
        &\quad |\hat{c}(h,U,I,S^m) - c(h,U,I)| > \epsilon \:\bigg|\: \\
        &\quad \forall h \in \mathcal{H}, \forall (U,I) \in C_h: \hat{n}(h,U,I,S^m) \geq l \biggr] \\
    &\leq \sum_{h \in \mathcal{H}} \sum_{(U,I) \in C_h}
    \Prob\biggl[ |\hat{c}(h,U,I,S^m) - c(h,U,I)| > \epsilon \:\bigg|\: \\ 
        &\quad \forall h \in \mathcal{H}, \forall (U,I) \in C_h: \hat{n}(h,U,I,S^m) \geq l \biggr] \\
    &= \sum_{h \in \mathcal{H}} \sum_{(U,I) \in C_h} \Prob\biggl[{|\hat{c}(h,U,I,S^m) - c(h,U,I)| > \epsilon} \:\bigg|\: \\
        &\quad {\hat{n}(h,U,I,S^m) \geq l}\biggr]
\end{align*}
By using the triangle inequality:
\begin{align*}
    &\sum_{h \in \mathcal{H}} \sum_{(U,I) \in C_h} \Prob\biggl[{|\hat{c}(h,U,I,S^m) - c(h,U,I)| > \epsilon} \:\bigg|\: \\
        &\quad {\hat{n}(h,U,I,S^m) \geq l}\biggr] \\
    &= \sum_{h \in \mathcal{H}} \sum_{(U,I) \in C_h} \Prob\biggl[ |\hat{\mu}_y(h,U,I,S^m) - \hat{\mu}_h(h,U,I,S^m) \\
        &\quad - \mu_y(h,U,I) + \mu_h(h,U,I)| > \epsilon \:\bigg|\: {\hat{n}(h,U,I,S^m) \geq l}\biggr] \\
    &\leq \sum_{h \in \mathcal{H}} \sum_{(U,I) \in C_h} \Prob\biggl[| \hat{\mu}_h(h,U,I,S^m) - \mu_h(h,U,I) | \\
    &\quad + | \mu_y(h,U,I) - \hat{\mu}_y(h,U,I,S^m) | > \epsilon \:\bigg|\:\\
    &\quad \hat{n}(h,U,I,S^m) \geq l \biggr]
\end{align*}
Since $a + b \geq \epsilon$ implies that either $a \geq \epsilon/2$ or $b \geq \epsilon/2$:
\begin{align*}
    &\sum_{h \in \mathcal{H}} \sum_{(U,I) \in C_h} \Prob\biggl[| \hat{\mu}_h(h,U,I,S^m) - \mu_h(h,U,I) | \\
        &\quad + | \mu_y(h,U,I) - \hat{\mu}_y(h,U,I,S^m) | > \epsilon \:\bigg|\: \\
        &\quad \hat{n}(h,U,I,S^m) \geq l \biggr] \\
    &\leq \sum_{h \in \mathcal{H}} \sum_{(U,I) \in C_h} \Prob\biggl[ | \hat{\mu}_h(h,U,I,S^m) - \mu_h(h,U,I) | > \frac{\epsilon}{2} \\
        &\quad \:\vee\: | \mu_y(h,U,I) - \hat{\mu}_y(h,U,I,S^m) | > \frac{\epsilon}{2} \:\bigg|\: \\
        &\quad \hat{n}(h,U,I,S^m) \geq l \biggr]
\end{align*}
And by using the union-bound once again:
\begin{align*}
    &\sum_{h \in \mathcal{H}} \sum_{(U,I) \in C_h} \Prob\biggl[ | \hat{\mu}_h(h,U,I,S^m) - \mu_h(h,U,I) | > \frac{\epsilon}{2} \\
        &\quad \:\vee\: | \mu_y(h,U,I) - \hat{\mu}_y(h,U,I,S^m) | > \frac{\epsilon}{2} \:\bigg|\: \\
        &\quad \hat{n}(h,U,I,S^m) \geq l \biggr] \\
    &\leq \sum_{h \in \mathcal{H}} \sum_{(U,I) \in C_h} \Prob\biggl[ | \hat{\mu}_h(h,U,I,S^m) - \mu_h(h,U,I) | > \frac{\epsilon}{2} \\
    &\quad\quad \:\bigg|\:  \hat{n}(h,U,I,S^m) \geq l \biggr] \\
    &\quad + \Prob\biggl[ | \mu_y(h,U,I) - \hat{\mu}_y(h,U,I,S^m) | > \frac{\epsilon}{2} \\
    &\quad\quad \:\bigg|\:  \hat{n}(h,U,I,S^m) \geq l  \biggr]
\end{align*}

We would like to use Chernoff inequality (Lemma \ref{abs_chernoff_thm}) to bound the probability with a confidence of $1 - \delta/2$. However,
in order to do so, we must fix the number of samples, $\hat{n}(h,U,I,S^m)$, that $h$ maps to a certain category (rather than using a random variable).
Note that for $\hat{n}(h,U,I,S^m) \geq l $ the probability is maximized at $\hat{n}(h,U,I,S^m) = l$, so we will assume that $\hat{n}(h,U,I,S^m) =l$.  We denote by $S^l|_{(h,U,I)}$ the sub-sample with $[x\in U,h(x)\in I]$, and its size is $l$. 


Now, in order to use Chernoff inequality, we define two random variables, $\hat{Z}_y(h,U,I)$ and $\hat{Z}_h(h,U,I)$, as follows:
\begin{align*}
    &\hat{Z}_y(h,U,I) := \frac{1}{l}\sum_{(x_i, y_i) \in S^l|_{(h,U,I)}}y_i \\
    &\hat{Z}_h(h,U,I) := \frac{1}{l}\sum_{(x_i, y_i) \in S^l|_{(h,U,I)}}h(x_i)
\end{align*}
and we observe that
\begin{align*}
     &\Ex\left[ \hat{Z}_y(h,U,I) \right] = \mu_h(h,U,I)\\
     &\Ex\left[ \hat{Z}_h(h,U,I) \right] = \mu_y(h,U,I)
\end{align*}


Using this notation,
\begin{align*}
    &\Prob\condsb{S^m \in B^m}{S^m \in G^{m,l}} \\
    &\leq \sum_{h \in \mathcal{H}}\sum_{(U,I) \in C_h}\biggl[ 
    \Prob\left[ \left| \hat{Z}_y(h,U,I) - \mu_h(h,U,I) \right| > \frac{\epsilon}{2} \right] \\
    &\quad + \Prob\left[ \left| \hat{Z}_h(h,U,I) - \mu_y(h,U,I) \right| > \frac{\epsilon}{2} \right]
    \biggr]
    \\
    &\leq \sum_{h \in \mathcal{H}} \sum_{(U,I) \in C_h}{ 4 e^{ - \frac{\epsilon^2}{2} l } }
    \leq \frac{4|\Gamma||\mathcal{H}|}{\lambda} e^{ - \frac{\epsilon^2}{2} l }
\end{align*}

We would like to set $l$ so that $\condprob{S^m \in B^m}{S^m \in G^{m,l}}$ will be at most $\delta/2$, as follows,
\begin{align*}
    &\frac{4|\Gamma||\mathcal{H}|}{\lambda} e^{ - \frac{\epsilon^2}{2} l } \leq \frac{\delta}{2}
    \iff 
    l \geq \frac{ 2 }{ \epsilon^2 } \log \left( \frac{ 8|\Gamma||\mathcal{H}| }{ \delta \lambda } \right)
    \\
\end{align*}
Hence we set
\[
l = \frac{ 2 }{ \epsilon^2 } \log \left( \frac{ 8|\Gamma||\mathcal{H}| }{ \delta \lambda } \right)
\]

Next, we will bound $\Prob\left[S^m \in \overline{G^{m,l}}\right]$ by $\delta/2$.

Since $m \geq m_{\mathcal{H}}(\epsilon, \delta, \psi, \gamma, \lambda)$ and since $p(h,U,I) \geq \gamma \psi$ for any $h \in \mathcal{H}$ and $(U,I) \in C_h$, we know that for any $h \in \mathcal{H}$ and $(U,I) \in C_h$:
\[
m \geq
\frac{4l}{\gamma \psi}
=
\frac{ 8 \log \left( \frac{ 8|\Gamma||\mathcal{H}| }{ \delta \lambda } \right) }{\epsilon^2 \gamma \psi}
\]

Thus, the expected number of samples we have in each interesting category, is at least twice the value of $l$,
\[
\Ex[\hat{n}(h,U,I,S)]
=
m p(h,U,I)
\geq
m \gamma \psi
\geq
2 l
\]

Thus, using the relative version of Chernoff bound, the upper bound we have on $l$, and the lower bound we have on $m$, for any $h \in \mathcal{H}$ and for any interesting category $(U,I) \in C_h$, the probability that $S^m$ has less than $l$ samples in the category $(U,I)$ is bounded by:
\begin{align*}
    &\Prob[\hat{n}(h,U,I,S) \leq l] \\
    &\leq \Prob\biggl[\hat{n}(h,U,I,S) \leq \frac{\Ex[\hat{n}(h,U,I,S)]}{2}\biggr] \\
    &\leq e^{-\frac{\Ex[\hat{n}(h,U,I,S)]}{8}} \leq \frac{\lambda \delta}{2|\Gamma||\mathcal{H}|}
\end{align*}

And, by using the union bound:
\begin{align*}
    &\Prob[S^m \in \overline{G^{m,l}}] \\
    &= \Prob\left[ \exists h \in \mathcal{H}, \exists (U,I) \in C_h : \hat{n}(h,U,I,S) < l \right] \\
    &\leq |C_h|\frac{\lambda \delta}{2|\Gamma|} \leq \frac{\delta}{2}
\end{align*}

Thus, overall:
\begin{align*}
    &\Prob[S^m \in B^m] \\
    &\leq \condprob{S^m \in B^m}{S^m \in G^{m,l}} \\
    &\quad + \Prob[S^m \in \overline{G^{m,l}}] \leq \delta/2 + \delta/2 = \delta
\end{align*}
as required.

\end{proof}
}









\section{Predictor Classes with Finite Graph Dimension}
\label{sec:Graph}
Throughout this section we assume that the predictions set $\mathcal{Y}$ is discrete.
This assumption allows us to analyze the multicalibration generalization of possibly infinite hypothesis classes with finite known multiclass complexity measures such as the graph-dimension.
(We discuss the case of $\mathcal{Y} = [0,1]$ at the end of the section.)
Recall that in this setup, the prediction-intervals set, $\Lambda$, contains singleton intervals with values taken from $\mathcal{Y}$, namely, $\Lambda = \left\{\{v\} \:|\: v \in \mathcal{Y}\right\}$.
Thus, if a prediction, $h(x)$ is in the interval $\{v\}$, it means the prediction value is exactly $v$, i.e., $h(x) \in \{v\}\Leftrightarrow h(x) = v$.
As we have mentioned earlier, part of our technique is to reduce multicalibration generalization to the generalization analysis of multiple binary hypothesis classes 
to get sample complexity bounds.
The Fundamental Theorem of Statistical Learning (see Theorem \ref{fund_thm}, Section \ref{app:usefulThms} in the Supplementary material) provides tight sample complexity bounds for uniform convergence for binary hypothesis classes. 
A direct corollary of this theorem indicates that by using ``large enough'' sample,
the difference between the true probability to receive a positive outcome and the estimated proportion of positive outcomes, is small, with high probability. 
\begin{corollary}\label{fund_cor}
Let $\mathcal{H} \subseteq \{0,1\}^\mathcal{X}$ be a binary hypothesis class with $VCdim(\mathcal{H}) \leq d$. Then, there exists a constant $C \in \mathbb{R}$ such that for any distribution $D$, and parameters $\epsilon,\delta \in (0,1)$, if $S= \{x_i,y_i\}_{i=1}^m$ is a sample of $m$ i.i.d. examples from $D$, and
$
m \geq C((d + \log(1/\delta))/\epsilon^2)
$
then with probability at least $1-\delta$,
$
\forall h \in \mathcal{H} : \left| \frac{1}{m} \sum_{i=1}^{m}{h(x_i) - \Prob_{x \sim D}[h(x) = 1]} \right| < \epsilon
$.

\end{corollary}

 Before we move on, we want to emphasize the main technical challenge in deriving generalization bounds for infinite predictor classes. Unlike PAC learning, in multicalibration learning the distribution over the domain is dependent on the predictors class.
Each 
pair of $h\in\HH, v\in\Y$ induce a distribution over the domain points $x$ such that $h(x)=v$.
As the number of predictors in the class is infinite, we cannot apply a simple union bound over the various induced distributions. This is a main challenge in our proof.
%
%
In order to utilize the existing theory about binary hypothesis classes we have to represent the calibration error in terms of binary predictors.
For this purpose, we define the notion of ``binary predictor class'', $\mathcal{H}_v \subseteq \{0,1\}^\mathcal{X}$, that depends on the original predictor class $\mathcal{H}$ and on a given prediction value $v \in \mathcal{Y}$.
Each binary predictor $h_v \in \mathcal{H}_v$ corresponds to a predictor $h\in \HH$ and value $v\in\Y$ and predicts $1$ on domain points $x$ if $h$ predicts $v$ on them (and $0$ otherwise).

\begin{definition}[Binary Predictor]\label{bin_pred_fun}
Let $h \in \mathcal{H}$ be a predictor and let $v \in \mathcal{Y}$ be a prediction value.
The {\em binary predictor} of $h$ and $v$, denoted $h_v(x)$, is the binary function that receives $x \in \mathcal{X}$ and outputs 1 iff $h(x) = v$, i.e.,
$h_{v}(x) =  \indicator{h(x) = v}$.
%
\label{bin_pred_class}
The {\em binary predictor class} w.r.t. the original predictor class $\mathcal{H}$ and value $v \in \mathcal{Y}$, denoted by $\mathcal{H}_v$, is defined as $\mathcal{H}_v = \left\{ h_v : h \in \mathcal{H} \right\}$.
\end{definition}
The definition of binary predictors alone is not sufficient since it ignores the outcomes $y \in \{0,1\}$. Thus, we define true positive function, $\phi_{h_v} \in \Phi_{\mathcal{H}_v}$, that corresponds to a binary predictor $h_v$, such that given a pair $(x \in \mathcal{X}, y \in \{0,1\})$, 
it outputs $1$ iff $h_v(x) = 1$ and $y = 1$.

\begin{definition}[True positive function]\label{bin_predout_fun}
Let $\mathcal{H}_v \subseteq \{0,1\}^{\mathcal{X}}$ be a binary predictor class and let $h_v \in \mathcal{H}_v$ be a binary predictor.
Then, the true positive function w.r.t. $h_v$ is 
$\phi_{h_v}(x,y) := \indicator{h_v(x) = 1, y = 1}$.
%
\label{bin_predout_class}
The {\em true positive class}   of $\mathcal{H}_v$, is defined $\Phi_{\mathcal{H}_v} := \left\{ \phi_{h_v} : h_v \in \mathcal{H}_v \right\}$.
\end{definition}
Using the above definitions we can re-write the calibration error as follows.
Let $I_v = \{v\}$ be a singleton interval. Then, the calibration error and the empirical calibration errors take the following forms:
$
    c(h,U,I_v) = \Ex_{D}\condsb{y}{x \in U, h(x)= v} - v  = \Prob_{(x,y) \sim D}\condsb{y = 1}{x \in U, h(x) = v} - v.
$
\begin{align*}
    \hat{c}(h,U,I_v,S) &= \sum_{i=1}^{m}\frac{\indicator{x_i \in U, h(x_i) = v}
    }{\sum_{j=1}^{m}{\indicator{x_j \in U, h(x_j) = v}}}y_i - v 
    = \frac{\sum_{i=1}^{m}{\indicator{x_i \in U, h(x_i) = v, y_i = 1}}} {\sum_{j=1}^{m}{\indicator{x_j \in U, h(x_j) = v}}} - v.
\end{align*}

The probability term in the calibration error notion is conditional on the subpopulation $U \in \Gamma$ and on the prediction value $h(x)$. Thus, different subpopulations and different predictors induce  different distributions on the domain $\mathcal{X}$.
To understand the challenge, consider the collection of conditional distributions induced by $h \in \mathcal{H}$ and an interesting category $(U,I)$. Since $\HH$ is infinite, we have an infinite collection of distributions, and guaranteeing uniform convergence for such a family of distributions is challenging.
In order to use the fundamental theorem of learning (Theorem \ref{fund_thm}), we circumvent this difficulty by re-writing the calibration error as follows.
\begin{align*}
    c(h,U,I_v) &= \Prob_{(x,y) \sim D}\condsb{y = 1}{x \in U, h(x) = v} - v = \frac{\Prob\condsb{y=1, h(x)=v}{x \in U} }{\Prob\condsb{h(x)=v}{x \in U}} - v.
\end{align*}
Later , we will separately approximate the numerator and denominator.\\
Finally, we use the definitions of binary predictor, $h_v$, and true positive functions $\phi_{h_v}$, to represent the calibration error in terms of binary functions. Thus, the calibration error and the empirical calibration error take the following forms:
\[
    c(h,U,I_v) = \frac{\Prob\condsb{y=1, h(x)=v}{x \in U} }{\Prob\condsb{h(x)=v}{x \in U}} - v = \frac{\Prob\condsb{\phi_{h_v}(x,y) = 1}{x \in U} }{\Prob\condsb{h_v(x) = 1}{x \in U}} - v,
\]
\[
    \hat{c}(h,U,I_v,S) = \frac{\sum_{i=1}^{m}{\indicator{x_i \in U, h(x_i) = v, y_i = 1}}} {\sum_{j=1}^{m}{\indicator{x_j \in U, h(x_j) = v}}} - v = \frac{\sum_{i=1}^{m}{\indicator{x_i \in U, \phi_{h_v}(x_i,y_i) = 1}}} {\sum_{j=1}^{m}{\indicator{x_j \in U, h_v(x_j) = 1}}} - v.
\]
Since the calibration error as written above depends on binary predictors, if we can prove that the complexity of the hypothesis classes containing them has finite VC-dimension, then we will be able to approximate for each term separately.
Recall that in this section we are dealing with multiclass predictors, which means that we must use multiclass complexity notion. 
We analyze the generalization of calibration by assuming that the predictor class $\mathcal{H}$ has a finite graph-dimension.
The following lemma states that a finite graph dimension of $\mathcal{H}$ implies finite VC-dimension of the binary prediction classes $\mathcal{H}_v$ for any $v \in \mathcal{Y}$. This result guarantees good approximation for the denominator term, $\Prob\condsb{h_v(x) = 1}{x \in U}$, in the calibration error. 
We remark that while the following lemma is also a direct corollary when considering graph dimension as a special case of Psi-dimension \cite{BenDavid1995}, for completeness,
we provide a simple proof for in the Supplementary material.  

\begin{lemma}\label{lma:graph}
Let $\mathcal{H} \subseteq \mathcal{Y}^\mathcal{X}$ be a predictor class such that $d_G(\mathcal{H}) \leq d$. Then, for any $v \in \mathcal{Y}$, $VCdim(\mathcal{H}_v) \leq d$.
\end{lemma}

\ignore{
\begin{proof}
Let us assume that $VCdim(\mathcal{H}_v) > d$ and let $S$ be a sample of size $d+1$ such that $\mathcal{H}_v$ shatters $S$.

Let us define the function $f:S \rightarrow \mathcal{Y}$ as:
\[
\forall x \in S: f(x) = v
\]

Let $T \subseteq S$ be an arbitrary subset of $S$.

By assuming that $\mathcal{H}_v$ shatters $S$ we know that there exists $h_v \in \mathcal{H}_v$ such that:
\[
\forall x \in S: h_v(x)=1 \iff x \in T
\]
This means that for the corresponding predictor $h \in \mathcal{H}$:
\[
\forall x \in S: h(x) = v = f(x) \iff x \in T
\]

Thus, using our definition of $f$,
\[
\forall T \subseteq S, \exists h \in \mathcal{H}, \forall x \in S: h(x) = f(x) \iff x \in T
\]
Which means that $S$ is G-shattered by $\mathcal{H}$.

But, since $|S| > d$, it is a contradiction to the assumption that $d_G(\mathcal{H}) \leq d$.
\end{proof}
}

In addition to the complexity bound of the binary predictor classes $\mathcal{H}_v$, we would like to derive a bound on the VC-dimension of the prediction-outcome classes $\Phi_{\mathcal{H}_v}$ which would enable a good approximation of the numerator term, $\Prob\condsb{\phi_{h_v}(x,y) = 1}{x \in U}$ in the calibration error. This bound is achieved by using the following lemma that indicates that the VC-dimension of $\Phi_{\mathcal{H}_v}$ is bounded by the VC-dimension of $\mathcal{H}_v$.

\begin{lemma}\label{lma:phi}
Let $\mathcal{H}_v \subseteq \{0,1\}^\mathcal{X}$ be a binary predictor class with $VCdim(\mathcal{H}_v) \leq d$, and let $\Phi_{\mathcal{H}_v}$ be the true positive class w.r.t. $\mathcal{H}_v$.
Then, $VCdim(\Phi_{\mathcal{H}_v}) \leq d$.
\end{lemma}

\ignore{
\begin{proof}
Assume that $VCdim(\Phi_{\mathcal{H}}) > d$ and let $S$ be a sample of $d+1$ individuals and outcomes shattered by $\Phi_{\mathcal{H}}$.

Note that $y=0$ implies that $\forall h \in \mathcal{H}, \forall x \in \mathcal{X}: \phi_{h}(x,y) = 0$. Thus, $\forall (x,y) \in S: y=1$ (otherwise $S$ cannot be shattered).

Let $S_x = \{x_j : (x_j, y_j) \in S\}$. Observe that when $y=1$, $\forall h \in \mathcal{H}, \forall x \in \mathcal{X}: \phi_{h}(x,1) = h(x)$. Thus, the fact that $S$ is shattered by $\Phi_{\mathcal{H}}$ implies that $S_x$ is shattered by $\mathcal{H}$.

But, $|S_x| = d+1$. Thus, we have a contradiction to the assumption that $VCdim(\Phi_{\mathcal{H}}) > d$.
\end{proof}
}

The fact that the VC-dimensions of $\mathcal{H}_v$ and $\Phi_{\mathcal{H}_v}$ are bounded enables to utilize the existing theory and derive sampling bounds for accurate approximations for the numerator and the denominator of the calibration error with high probability, respectively. 
Lemma \ref{lma:numDenum} formalizes these ideas.


\ignore{
\begin{proof}
Let $\mathcal{H}_v$ be the binary prediction class of $\mathcal{H}$ (see \ref{bin_pred_class}).

Using Lemma \ref{lma:graph}, and since $d_G(\mathcal{H}) \leq d$, we know that $VCdim(\mathcal{H}_v) \leq d$.

In addition, note that:
\begin{align*}
    &\left| \frac{1}{m} \sum_{i=1}^{m}{\indicator{h(x_i) = v}} - \Prob_{x \sim D_U}[h(x) = v] \right| \\
        &\quad = \left| \frac{1}{m} \sum_{i=1}^{m}{h_v(x_i)} - \Prob_{x \sim D_U}[h_v(x) = 1] \right|
\end{align*}

Thus, the lemma follows directly from Corollary \ref{fund_cor}.
\end{proof}
}

\begin{lemma}\label{lma:numDenum}
Let $\mathcal{H} \subseteq \mathcal{Y}^\mathcal{X}$ be a predictor class with $d_G(\mathcal{H}) \leq d$. 
Let $v \in \mathcal{Y}$ be a prediction value and let $U \subset \mathcal{X}$ be a subpopulation. 
Then, there exist a constant $C\in \mathbb{R}$ such that for any distribution $D$ over $\X \times \{0,1\}$ and $\epsilon, \delta \in (0,1)$, if $D_U$ is the induced distribution on $U \times \{0,1\}$ and $S = \{x_i,y_i\}_{i=1}^m$ is a random sample of size $m \geq C \frac{d + \log(1/\delta)}{\epsilon^2}$ drawn i.i.d. according to $D_U$,
then with probability at least $1 - \delta$ for every $h\in \mathcal{H}$:
\[
\left| 
\frac{1}{m} \sum_{i=1}^{m}{\indicator{h(x_i)=v}} - \Prob_{
D_U}[h(x)=v] 
\right|
,
    \biggl| \frac{1}{m} \sum_{i=1}^{m}{\indicator{h(x_i)=v, y=1}}
    - \Prob_{
    D_U}[h(x)=1, y=1] \biggr| \leq \epsilon.
\]
\end{lemma}
\ignore{
\begin{proof}
Let $\mathcal{H}_v$ and $\Phi_{\mathcal{H}_v}$ be the binary prediction and binary prediction-outcome classes of $\mathcal{H}$.

Using Lemmas \ref{lma:graph} and \ref{lma:phi}, and since $d_G(\mathcal{H}) \leq d$, we know that $VCdim(\Phi_{\mathcal{H}_v}) \leq VCdim(\mathcal{H}_v) \leq d$.

In addition, note that:
\begin{align*}
    &\biggl| \frac{1}{m} \sum_{i=1}^{m}{\indicator{h(x_i) = v, y=1}} \\
        &\quad\quad - \Prob_{(x,y) \sim D_U}[h(x) = v, y=1] \biggr| \\
    &= \left| \frac{1}{m}  \sum_{i=1}^{m}{\phi_{h,v}(x_i,y_1)} - \Prob_{(x,y) \sim D_U}[\phi_{h,v}(x,y)] \right|
\end{align*}
and the lemma follows directly from Corollary \ref{fund_cor}.
\end{proof}
}
Having an accurate approximation of the denominator and numerator terms of the calibration error does not automatically implies good approximation for it. 
For example, any approximation error in the numerator is scaled by $1$ divided by the denominator's value. 
The following lemma tells us how accurate the approximations of the numerator and the denominator
should be in order to achieve good approximation of the entire fraction, given a lower bound on the true value of the denominator.
\begin{lemma}\label{lma:err}
Let $p_1, p_2, \tilde{p}_1, \tilde{p}_2, \epsilon, \psi \in [0,1]$ such that
$    p_1, \psi \leq p_2$ and $
    \left|p_1 - \tilde{p}_1\right|, \left|p_2 - \tilde{p}_2\right| 
    \leq \psi\epsilon/3$.
Then,
$\left| p_1/p_2 -\tilde{p}_1/\tilde{p}_2 \right| 
\leq \epsilon$.
\end{lemma}
\ignore{
\begin{proof}
Let us denote $\xi := \psi\epsilon/3$
\begin{align*}
    \frac{p_1}{p_2} - \frac{\tilde{p}_1}{\tilde{p}_2} &\leq \frac{p_1}{p_2} - \frac{p_1 - \xi}{p_2 + \xi} \\
    &= \frac{p_1(1 + \xi/p_2)}{p_2(1 + \xi/p_2)} - \frac{p_1 - \xi}{p_2(1 + \xi/p_2)} \\
    &= \frac{\xi}{p_2(1 + \xi/p_2)} \left[ \frac{p_1}{p_2} + 1 \right]
\end{align*}

Since $p_1, \psi \leq p_2$,
\begin{align*}
    &\frac{\xi}{p_2(1 + \xi/p_2)} \left[ \frac{p_1}{p_2} + 1 \right]
    \\
    &\leq \frac{\xi}{p_2} \left[ \frac{p_2}{\psi} + \frac{p_2}{\psi} \right]
    = \frac{2\xi}{\psi}
    \leq \frac{3\xi}{\psi}
    = \epsilon
\end{align*}

Similarly,
\begin{align*}
    \frac{\tilde{p}_1}{\tilde{p}_2} - \frac{p_1}{p_2} &\leq \frac{p_1 + \xi}{p_2 - \xi} - \frac{p_1}{p_2} \\
    &= \frac{p_1 + \xi}{p_2(1 - \xi/p_2)} - \frac{p_1(1 - \xi/p_2)}{p_2(1 - \xi/p_2)} \\
    &= \frac{\xi}{p_2(1 - \xi/p_2)} \left[ 1 + \frac{p_1}{p_2} \right] \\
\end{align*}

Since $p_1, \psi \leq p_2$,
\begin{align*}
    &\frac{\xi}{p_2(1 - \xi/p_2)} \left[ 1 + \frac{p_1}{p_2} \right] \\
    &\leq \frac{\xi}{p_2(1 - \xi/\psi)} \left[ \frac{p_2}{\psi} + \frac{p_2}{\psi} \right] \\
    &= \frac{2\xi}{\psi(1 - \xi/\psi)} = \frac{2\epsilon}{3(1 - \epsilon/3)} \\
    &\leq \frac{2\epsilon}{3(1 - 1/3)} = \epsilon
\end{align*}

Thus,
\[
\left| \frac{p_1}{p_2} - \frac{\tilde{p}_1}{\tilde{p}_2} \right| \leq \epsilon
\]
\end{proof}
}
%
Since multicalibration uniform convergence requires empirical calibration errors of interesting categories to be close to their respective (true) calibration errors, a necessary condition is to have a large sample from every large subpopulation $U \in \Gamma$.
The following lemma indicates the sufficient 
 sample size 
 to achieve a large subsample from every large subpopulation with high probability.
\begin{lemma} \label{lma:u_size}
Let $\gamma \in (0,1)$ and let $\Gamma_\gamma = \{U \in \Gamma \:|\: \Prob_{x \sim D}[x \in U] \geq \gamma\}$ be the collection of subpopulations from $\Gamma$ that has probability at least $\gamma$ according to $D$.
Let $\delta \in (0,1)$ and let $S = \{(x_i,y_i)\}_{i=1}^m$ be a random sample of $m$ i.i.d. examples from $D$.
Then, with probability at least $1 - \delta$, if
$
m \geq \frac{8}{\gamma} \log\left(|\Gamma|/\delta\right)
$, it holds that $\forall U \in \Gamma_\gamma: |S \cap U| > \frac{\gamma m}{2}$.
\end{lemma}

\ignore{
\begin{proof}
Let $\text{P}_U$ denote the probability of subpopulation $U$:
\[
\text{P}_U := \Prob_{x \sim D}\left[x \in U\right]
\]

Using the relative Chernoff bound (Lemma \ref{rel_chernoff_thm}) and since $\Ex[|S \cap U|] = m \text{P}_U$, we can bound the probability of having a small sample size in $U$. Namely, if $\text{P}_U \geq \gamma$, then:
\begin{align*}
    \Prob_D\left[|S \cap U| \leq \frac{\gamma m }{2}\right] &\leq \Prob_D\left[|S \cap U| \leq \frac{m \text{P}_U }{2}\right] \\
        &\leq e^{-\frac{m \text{P}_U}{8}} \leq e^{-\frac{\gamma m}{8}}
\end{align*}

Thus, for any $U \in \Gamma_\gamma$, if $m \geq \frac{8 \log\left(\frac{|\Gamma|}{\delta}\right)}{\gamma}$, 
then, with probability of at least $1 - \frac{\delta}{|\Gamma|}$,
\[
|S \cap U| > \frac{\gamma m}{2}
\]

Finally, using the union bound, with probability at least $1 - \delta$, for all $U \in \Gamma_\gamma$,
\[
|S \cap U| > \frac{\gamma m}{2}
\]
\end{proof}
}
The following theorem combines all the intuition described above and prove an upper bound on the sample size needed to achieve multicalibration uniform convergence. It assumes that the predictor class $\mathcal{H}$ has a finite graph-dimension, $d_G(\mathcal{H})$ and uses Lemma \ref{lma:graph} and Lemma \ref{lma:phi} to derive an upper bound on the VC-dimension of $\mathcal{H}_v$ and $\Phi_{\mathcal{H}_v}$. Then, it uses Lemma \ref{lma:numDenum} to bound the sample complexity for ``good'' approximation of the numerator and the denominator of the calibration error.

\graphThm*
%
The proof of Theorem \ref{main:thm:graph_ub} uses the relative Chernoff bound (Lemma \ref{rel_chernoff_thm}) to show that with probability at least $1-\delta/2$, every subpopulation $U \in \Gamma$ with $\Prob_D[U] \geq \gamma$, has a sub-sample of size at least $\frac{\gamma m}{2}$, namely $|S \cap U| \geq \frac{\gamma m}{2}$. 
Then, it uses Lemmas \ref{lma:graph} and \ref{lma:phi} to show that for every $v \in \mathcal{Y}$, $VCdim(\Phi_{\mathcal{H}_v}) \leq VCdim(\mathcal{H}_v) \leq d_G(\mathcal{H})$. 
It proceeds by applying Lemma \ref{lma:numDenum} to show that, with probability at least $1-\delta/2$, for every prediction value $v \in \mathcal{Y}$ and every subpopulation $U \in \Gamma$, if $|S \cap U| \geq \frac{\gamma m}{2}$, then:
$\biggl|\Prob\condsb{\phi_{h_v}(x,y) = 1}{x \in U}
     - \frac{1}{|S \cap U|}\sum_{i=1}^{m}{\indicator{x_i \in U, \phi_{h_v}(x_i,y_i) = 1}}\biggr| \leq \frac{\psi\epsilon}{3}$, 
     and $\biggl|\Prob\condsb{h_v(x) = 1}{x \in U} 
    - \frac{1}{|S \cap U|}\sum_{j=1}^{m}{\indicator{x_j \in U, h_v(x_j) = 1}}\biggr| \leq \frac{\psi\epsilon}{3}$.\\
Finally, it concludes the proof of Theorem \ref{main:thm:graph_ub} using Lemma \ref{lma:err}.
\ignore{
\begin{proof}
Let $S = \{(x_1, y_1), ..., (x_m, y_m)\}$ be a sample of $m$ labeled examples drawn i.i.d. according to $D$, and let $S_U := \{(x,y) \in S : x \in U\}$ be the samples in $S$ that belong to subpopulation $U$.

Let $\Gamma_\gamma$ denote the set of all subpopulations $U \in \Gamma$ that has probability of at least $\gamma$:
\[
\Gamma_\gamma := \{U \in \Gamma \:|\: \Prob_{x \sim D}[x \in U] \geq \gamma\}
\]

%



Let us assume the following lower bound on the sample size:
\[
m \geq \frac{8 \log\left( \frac{2 |\Gamma|}{\delta} \right)}{\gamma}
\]

Thus, using Lemma \ref{lma:u_size},
we can bound the probability of having a subpopulation $U \in \Gamma_\gamma$ with small number of samples.
Namely, we know that with probability of at least $1 - \delta/2$, for every $U \in \Gamma_\gamma$:
\[
|S_U| \geq \frac{\gamma m}{2}
\]

Next, we would like to show that having a large sample size in $U$ implies accurate approximation of the calibration error, with high probability, for any interesting category in $U$. For this purpose, let us define $\epsilon', \delta'$ as:
\begin{align*}
    &\epsilon' := \frac{\psi \epsilon}{3}
    &\delta' := \frac{\delta}{4|\Gamma||\mathcal{Y}|}
\end{align*}

By using Lemma \ref{lma:numDenum} and since $d_G(\mathcal{H}) \leq d$, we know that there exists some constant $a > 0$, such that, for any $v \in \mathcal{Y}$ and any $U \in \Gamma_\gamma$, with probability at least $1 - \delta'$, a random sample of $m_1 \geq m_1^{UC}(\epsilon', \delta')$ examples from $U$, where,
\[
m_1^{UC}(\epsilon', \delta') 
\leq
a \frac{d + \log(1/\delta')}{\epsilon'^2}
=
9a \frac{d + \log(\frac{4|\Gamma||\mathcal{Y}|}{\delta})}{\epsilon^2 \psi^2}
\]
will have,
\begin{align*}
    &\forall h \in \mathcal{H}: \biggl| \frac{1}{m_1}\sum_{x' \in S_U}{\indicator{h(x') = v}} \\
        &\quad - \Prob\condsb{h(x) = v}{x \in U} \biggr| \leq  \epsilon' = \frac{\psi \epsilon}{3}
\end{align*}

By using Lemma \ref{lma:numDenum} and since $d_G(\mathcal{H}) \leq d$, we know that for any $v \in \mathcal{Y}$ and any $U \in \Gamma_\gamma$, with probability at least $1 - \delta'$, a random sample of $m_2 \geq m_2^{UC}(\epsilon', \delta')$ labeled examples from $U \times \{0,1\}$, where,
\[
m_2^{UC}(\epsilon', \delta') 
\leq
a \frac{d + \log(1/\delta')}{\epsilon'^2}
=
9a \frac{d + \log(\frac{4|\Gamma||\mathcal{Y}|}{\delta})}{\epsilon^2 \psi^2}
\]
will have,
\begin{align*}
    &\forall h \in \mathcal{H}: 
    \biggl| \frac{1}{m_2}\sum_{(x', y') \in S_U}{\indicator{h(x') = v, y' = 1}} \\
        &\quad - \Prob\condsb{h(x)=v, y=1}{x \in U} \biggr| \leq  \epsilon' = \frac{\psi \epsilon}{3}
\end{align*}

Let us define the constant $a'$ in a manner that sets an upper bound on both $m_1$ and $m_2$:
\[
a' := 18 a
\]
and let $m'$ be that upper bound:
\[
m' 
:= 
a' 
\frac{d + \log\left(\frac{|\Gamma||\mathcal{Y}|}{\delta}\right)}
{\psi^2 \epsilon^2}
\geq
m_1, m_2
\]

Then, by the union bound, if for all subpopulations $U \in \Gamma_\gamma$, $|S_U| \geq m'$, then, with probability at least $1 - 2|\Gamma||\mathcal{Y}|{}\delta' = 1 - \frac{\delta}{2}$:
\begin{align*}
    &\forall h \in \mathcal{H}, \forall U \in \Gamma_\gamma, \forall v \in \mathcal{Y}: \\
        &\quad \biggl| \frac{1}{|S_U|}\sum_{(x',y') \in S_U}{\indicator{h(x') = v}} \\
            &\quad\quad - \Prob\condsb{h(x) = v}{x \in U} \biggr| \leq \frac{\psi \epsilon}{3} \\
    &\forall h \in \mathcal{H}, \forall U \in \Gamma_\gamma, \forall v \in \mathcal{Y}: \\
        &\quad \biggl| 
        \frac{1}{|S_U|}\sum_{(x', y') \in S_U}{\indicator{h(x')=v, y'=1}} \\
        &\quad - \Prob\condsb{h(x) = v, y=1}{x \in U} \biggr| \leq  \frac{\psi \epsilon}{3}
\end{align*}

Let us choose the sample size $m$ as follows:
\[
m
:=
\frac{2m'}{\gamma}
=
2a 
\frac{d + \log\left(\frac{|\Gamma||\mathcal{Y}|}{\delta}\right)}
{\psi^2 \epsilon^2 \gamma}
\]

Recall that with probability at least $1 - \delta/2$, for every $U \in \Gamma_\gamma$: 
\[
|S_U| \geq \frac{\gamma m}{2} = m'
\]

Thus, using the union bound once again, with probability at least $1 - \delta$: 
\begin{align*}
    &\forall h \in \mathcal{H}, \forall U \in \Gamma_\gamma, \forall v \in \mathcal{Y}: \\
    &\quad  \biggl|  \frac{1}{|S_U|}\sum_{x' \in S_U}{\indicator{h(x') = v}} \\
    &\quad\quad - \Prob\condsb{h(x) = v}{x \in U} \biggr| \leq \frac{\psi \epsilon}{3}
    \\
    &\forall h \in \mathcal{H}, \forall U \in \Gamma_\gamma, \forall v \in \mathcal{Y}: \\
    &\quad \biggl| \frac{1}{|S_U|}\sum_{(x',y') \in S_U}{\indicator{h(x')=v, y'=1}} \\
    &\quad\quad - \Prob\condsb{h(x) = v, y = 1}{x \in U} \biggr| \leq \frac{\psi \epsilon}{3}
\end{align*}

To conclude the theorem, we need show that having $\psi\epsilon/3$ approximation to the terms described above, implies accurate approximation to the calibration error.
For this purpose, let us denote:
\begin{align*}
    &p_1(h,U,v) := \Prob\condsb{h(x) = v, y = 1}{x \in U} 
    \\
    &p_2(h,U,v) := \Prob\condsb{h(x) = v}{x \in U} 
    \\
    &\tilde{p}_1(h,U,v) := \frac{1}{|S_U|}\sum_{(x', y') \in S_U}{\indicator{h(x')=v, y'=1}} 
    \\
    &\tilde{p}_2(h,U,v) := \frac{1}{|S_U|}\sum_{x' \in S_U}{\indicator{h(x') = v}}
\end{align*}

Then, with probability at least $1 - \delta$:
\begin{align*}
    &\forall h \in \mathcal{H}, \forall U \in \Gamma_\gamma, \forall v \in \mathcal{Y}: \\
    &\quad \biggl| \tilde{p}_2(h,U,v) - p_2(h,U,v) \biggr| \leq \frac{\psi \epsilon}{3}
    \\
    &\forall h \in \mathcal{H}, \forall U \in \Gamma_\gamma, \forall v \in \mathcal{Y}: \\
    &\quad \biggl| \tilde{p}_1(h,U,v) - p_1(h,U,v) \biggr| \leq \frac{\psi \epsilon}{3}
\end{align*}


Using Lemma \ref{lma:err}, 
for all $h \in \mathcal{H}$, $U \in \Gamma_\gamma$ and $v \in \mathcal{Y}$,
if $p_2(h,U,v) \geq \psi$,
then:
\[
\left| 
\frac{p_1(h,U,v)}{p_2(h,U,v)} - 
\frac{\tilde{p}_1(h,U,v)}{\tilde{p}_2(h,U,v)} 
\right|
\leq 
\epsilon
\]

Thus, since
\begin{align*}
    & c(h,U,\{v\}) = \frac{p_1(h,U,v)}{p_2(h,U,v)} \\
    & \hat{c}(h,U,\{v\},S) = \frac{\tilde{p}_1(h,U,v)}{\tilde{p}_2(h,U,v)}
\end{align*}
then with probability at least $1 - \delta$:
\begin{align*}
    &\forall h \in \mathcal{H}, \forall U \in \Gamma, \forall v \in \mathcal{Y}: \\
    &\quad \Prob[x \in U] \geq \gamma, \Prob\condsb{h(x) = v}{x \in U} \geq \psi \Rightarrow \\
    &\quad\quad \left| c(h,U,\{v\}) - \hat{c}(h,U,\{v\},S) \right| \leq \epsilon
\end{align*}
\end{proof}
}
Similarly to the discussion of Corollary \ref{cor:finite}, we derive the following corollary from Theorem \ref{main:thm:graph_ub}.
\begin{corollary}
Let $\mathcal{H} \subseteq \mathcal{Y}^\mathcal{X}$ be a predictor class with $d_G(\mathcal{H}) \leq d$ and let $D$ be a distribution over $\mathcal{X} \times \{0,1\}$.
Let $S$ be a random sample of $m$ examples drawn i.i.d. from $D$ and let $h \in \mathcal{H}$ be $(\alpha, \gamma, \psi)$-- empirically multicalibrated predictor on $S$.
Then, there exists a constant $C > 0$ such that for any $\epsilon, \delta \in (0,1)$, if
$m \geq \frac{C}{\epsilon^2 \psi^2 \gamma}\left(d + \log\left(|\Gamma||\mathcal{Y}|/\delta\right)\right)$,  
then with probability at least $1 - \delta$, $h$ is $(\alpha + \epsilon, \gamma, \psi)$--multicalibrated w.r.t. the underlying distribution $D$.
\end{corollary}
\noindent{\bf Finite versus continuous $\Y$:}
We have presented all the results for the infinite predictor class using a finite prediction-interval set  $\Lambda=\{\{v\}|v\in \Y\}$. We can extend our results to the continuous $\Y=[0,1]$ in a straightforward way. We can simply round the predictions to a value $j\lambda$, and there are $1/\lambda$ such values. This will result in an increase in the calibration error of at most $\lambda$. (Note that in the finite predictor class case, we have a more refine analysis that does not increase the calibration error by $\lambda$.) The main issue with this approach is that the graph-dimension depends on the parameter $\lambda$ through the induced values $j\lambda$. Since we select $\lambda$ and the points $j\lambda$, the magnitude of graph-dimension depends not only on the predictor class but also on parameters which are in our control, and therefore harder to interpret. For this reason we preferred to present our results for the finite $\Y$ case, and remark that one can extend them to the continuous $\Y=[0,1]$ case.

\section{Lower Bound}
\label{sec:lowerBound}
We prove a lower bound for the required number of samples to get multicalibration uniform convergence. The proof is done by considering a predictor class with a single predictor that maps $\gamma\psi$ fraction of the population to $1/2 +\epsilon$. We show that this class has multicalibration uniform convergence property for $1/2+\epsilon$ and then show how to use this property to distinguish between biased coins, which yield a lower bound of $\Omega(\frac{1}{\psi\gamma\epsilon^2}\ln (1/\delta))$ on the sample complexity.
\lowerBoundThm*

\section{Discussion and Future Work}

In this work, we derived uniform convergence for multicalibration notion. 
We provided upper and lower bounds on the sample size needed to guarantee uniform convergence of multicalibration for both finite and infinite predictor classes. For infinite classes, the bounds based on the graph dimension of the class. 

While our upper bounds depend logarithmically on the size of the predictor class (or on the graph dimension for infinite predictor classes), out lower bounds do not match them.
We believe, in general, that dependence of $\log(|\mathcal{H}|)$ is essential for the sample complexity, similar to lower bounds on sample complexity for agnostic PAC learning. 
Future work is needed to better understand the relation between multicalibration uniform convergence and the complexity of the predictor class.

Another interesting problem is to enable an infinite number of subpopulations defined by  a class of binary functions with bounded VC-dimension. Deriving uniform convergence bounds in this setting will require overcoming some new challenges, since one cannot simply enumerate all subpopulations.

\section*{Acknowledgments}
\label{sec:acknowledgments}
This project has received funding from the European Research Council (ERC) under the European Union’s Horizon 2020 research and innovation program (grant agreement No. 882396), and by the Israel Science Foundation (grant number 993/17).
Lee Cohen is a fellow of the Ariane de Rothschild Women Doctoral Program.

\bibliographystyle{apa-good}
\bibliography{refs}

\newpage
\appendix

\section{Useful Definitions \& Theorems}\label{app:usefulThms}
Throughout this paper, we use the following standard Chernoff bounds.
\begin{lemma}[Absolute Chernoff Bound]\label{abs_chernoff_thm}
Let $X_1, ..., X_n$ be i.i.d. binary random variables with $\Ex[X_i] = \mu$ for all $i \in [n]$. Then, for any $\epsilon > 0$:
$\Prob\left[ \left| \frac{1}{n}\sum_{i=1}^{n}{X_i} - \mu \right| \geq \epsilon \right]
\leq
2\exp(-2 \epsilon^2 n)$.
\end{lemma}
\begin{lemma}[Relative Chernoff Bound]\label{rel_chernoff_thm}
Let $X_1, ..., X_n$ be i.i.d. binary random variables and let $X$ denote their sum. Then, for any $\epsilon \in (0,1)$:$
\Prob\left[ X \leq (1-\epsilon)\Ex[X] \right]
\leq
\exp(-\epsilon^2 \Ex[X]/2)$.
\end{lemma}

Next, the definition of Vapnik–Chervonenkis dimension, following by Uniform convergence for statistical learning and the Fundamental Theorem of Statistical Learning.
\begin{definition}\label{def:vc}[VC-dimension]
Let $\mathcal{\mathcal{H}} \subseteq \{0,1\}^\mathcal{X}$ be a hypothesis class. A subset $S = \{x_1, ..., x_{|S|}\} \subseteq \mathcal{X}$ is shattered by $\mathcal{H}$ if:
$\left| \left\{
\left(h(x_1), ..., h(x_{|S|})\right) : h \in \mathcal{H}
\right\} \right|
= 2^{|S|}$.
The VC-dimension of $\mathcal{H}$, denoted $VCdim(\mathcal{H})$, is the maximal cardinality of a subset $S \subseteq \mathcal{X}$ shattered by $\mathcal{H}$.
\end{definition}

\begin{definition}[Uniform convergence for statistical learning]
Let $\mathcal{H} \subseteq \mathcal{Y}^\mathcal{X}$ be a hypothesis class.
We say that $\mathcal{H}$ has the uniform convergence property w.r.t. loss function $\ell$ if there exists a function $m_{\mathcal{H}}^{sl}(\epsilon, \delta) \in \mathbb{N}$ such that for every $\epsilon, \delta \in (0,1)$ and for every probability distribution $D$ over $\mathcal{X} \times \{0,1\}$, if $S$ is a sample of $m \geq m_\mathcal{H}^{sl}(\epsilon, \delta)$ examples drawn i.i.d. from to $D$, then, with probability of at least $1 - \delta$,
for every $h \in \mathcal{H}$, the difference between the risk and the  empirical risk is at most $\epsilon$. Namely, with probability $1-\delta$,
$\forall h \in \mathcal{H}: \left|L_S(h) - L_D(h)\right| \leq \epsilon$.
\end{definition}

\begin{theorem}\label{fund_thm}[The Fundamental Theorem of Statistical Learning]
Let $\mathcal{H} \subseteq \{0,1\}^\mathcal{X}$ be a binary hypothesis class  with $VCdim(\mathcal{H}) = d$ and let the loss function, $\ell$, be the $0-1$ loss.
Then, 
$\mathcal{H}$ has the uniform convergence property with sample complexity $m_\mathcal{H}^{UC}(\epsilon, \delta)=
\Theta\left(\frac{1}{\epsilon^2}\left(d + \log(1/\delta)\right)\right)$.
\end{theorem}

\section{Proofs for Section \ref{sec:Finite}}
\begin{proof}(Proof of Theorem \ref{main:thm:finite_gen_thm})
\\
Let $S^m = \{(x_1, y_1), ..., (x_m, y_m)\}$ be a random sample of size $m \geq m_{\mathcal{H}}(\epsilon, \delta, \psi, \gamma, \lambda)$ labeled examples drawn i.i.d. according to $D$.

For convenience, throughout the proof we use the following notations. We first define the quantities with respect to the distribution. For a given hypothesis $h\in H$,  group $U\in \Gamma$ and interval $I\in \Lambda$, we are interested in the subpoppulation which belongs to $U$ and for which $h$ prediction is in $I$, i.e.,  $[x\in U,h(x)\in I]$.
For this subpoppulation we define:
$p(h,U,I)$ the probability of being in this subpopulation, $\mu_y(h,U,I)$ the average $y$ value in the subpoppulation, and $\mu_h(h,U,I)$, the average prediction, i.e., $h(x)$. The three measures are with respect to the true distribution $D$.
Formally,
\begin{align*}
    & p(h,U,I) := \Prob_{D}[x \in U, h(x) \in I] \\
    & \mu_y(h,U,I) := \Ex_D\condsb{y}{x \in U, h(x) \in I} \\
    & \mu_h(h,U,I) := \Ex_{D}\condsb{h(x)}{x \in U, h(x) \in I}
\end{align*}
Similarly we denote the three empirical quantities with respect to the sample. Namely, we denote by $\hat{n}(h,U,I,S)$, $\hat{\mu}_y(h,U,I,S)$ and $\hat{\mu}_h(h,U,I,S)$ the number of samples, empirical outcome and empirical prediction, of the subpoppulation $[x\in U,h(x)\in I]$.
Formally,
\begin{align*}
    & \hat{n}(h,U,I,S) := \sum_{i=1}^{m}{\indicator{x_i \in U, h(x_i) \in I}} \\
    & \hat{\mu}_y(h,U,I,S) := \sum_{i=1}^{m}\frac{{\indicator{x_i \in U, h(x_i) \in I} }}{\hat{n}(h,I,U,S)} y_i \\
    & \hat{\mu}_h(h,U,I,S) := \sum_{i=1}^{m} \frac{{\indicator{x_i \in U, h(x_i) \in I} }}{\hat{n}(h,I,U,S)}h(x_i)
\end{align*}

Then, the calibration error and the empirical calibration error can be expressed as:
\begin{align*}
    & c(h,U,I) = \mu_y(h,U,I) - \mu_h(h,U,I) \\
    & \hat{c}(h,U,I,S) = \hat{\mu}_y(h,U,I,S) - \hat{\mu}_h(h,U,I,S)
\end{align*}

Let $C_h$ denote the collection of all interesting categories according to predictor $h$, namely,
\begin{align*}
    C_h := &\biggl\{(U,I) : U \in \Gamma, I \in \Lambda, \Prob_{D}[x \in U] \geq \gamma, \Prob_{D}\condsb{h(x) \in I}{x \in U} \geq \psi \biggr\}
\end{align*}

Note that every interesting category $(U,I) \in C_h$ has a probability of at least $\gamma \psi$, namely, for every $h \in \mathcal{H}$ and for any interesting category $(U,I) \in C_h$:
\begin{align*}
    &\Prob_{x \sim D}[x \in U, h(x) \in I] = \Prob_{x \sim D}\condsb{h(x) \in I}{x \in U} \cdot \Prob_{x \sim D}[x \in U] \geq \gamma \psi
\end{align*}

We define a ``bad'' event $B^m$ over the samples, as the event there exist some predictor and some interesting category for which the generalization error is larger than $\epsilon$.
\begin{align*}
    B^m := & \biggl\{ S \in (\mathcal{X} \times \{0,1\})^m : \exists h \in \mathcal{H} ,\exists (U,I) \in C_h : |\hat{c}(h,U,I,S) - c(h,U,I)| > \epsilon \biggr\}
\end{align*}

Bounding the probability that $S^m \in B^m$ by $\delta$ implies the theorem.
In order to do so, we would like to have a ``large enough'' induced sample in every interesting category.
For this purpose, we define the ``good'' event, $G^{m,l}$, as the event that indicates that for every predictor, each interesting category has at least $l$ samples.
\begin{align*}
    G^{m,l} := &\biggl\{ S \in (\mathcal{X} \times \{0,1\})^m : \forall h \in \mathcal{H}, \forall (U,I) \in C_h: \hat{n}(h,U,I,S) \geq l \biggr\}
\end{align*}

We will later set $l$ to achieve $\epsilon$-accurate approximation with confidence $\delta$ later. 
Note that $G^{m,l}$ is not the complement of $B^m$.

According to the law of total probability the following holds:
\begin{align*}
    \Prob[B^m] &= \Prob\condsb{B^m}{G^{m,l}}\Prob\left[G^{m,l}\right] + \Prob\condsb{B^m}{\overline{G^{m,l}}} \Prob\left[\overline{G^{m,l}}\right] \\
    &\leq  \Prob\condsb{B^m}{G^{m,l}} + \Prob\left[\overline{G^{m,l}}\right]
\end{align*}
We would like to bound each of the probabilities
$\condprob{B^m}{G^{m,l}}$ and $\Prob[\overline{G^{m,l}}]$ by $\delta/2$, in order to bound the probability of $B^m$ by $\delta$.
We start by bounding $\Prob\condsb{S^m \in B^m}{S^m \in G^{m,l}}$.
By using the union bound:
\begin{align*}
    &\Prob\condsb{S^m \in B^m}{S^m \in G^{m,l}} \\
    &= \Prob\biggl[ \exists h \in \mathcal{H} ,\exists (U,I) \in C_h : |\hat{c}(h,U,I,S^m) - c(h,U,I)| > \epsilon \:\bigg|\: \forall h \in \mathcal{H}, \forall (U,I) \in C_h: \hat{n}(h,U,I,S^m) \geq l \biggr] \\
    &\leq \sum_{h \in \mathcal{H}} \sum_{(U,I) \in C_h}
    \Prob\biggl[ |\hat{c}(h,U,I,S^m) - c(h,U,I)| > \epsilon \:\bigg|\: \forall h \in \mathcal{H}, \forall (U,I) \in C_h: \hat{n}(h,U,I,S^m) \geq l \biggr] \\
    &= \sum_{h \in \mathcal{H}} \sum_{(U,I) \in C_h} \Prob\biggl[{|\hat{c}(h,U,I,S^m) - c(h,U,I)| > \epsilon} \:\bigg|\: {\hat{n}(h,U,I,S^m) \geq l}\biggr]
\end{align*}
By using the triangle inequality:
\begin{align*}
    &\sum_{h \in \mathcal{H}} \sum_{(U,I) \in C_h} \Prob\biggl[{|\hat{c}(h,U,I,S^m) - c(h,U,I)| > \epsilon} \:\bigg|\: {\hat{n}(h,U,I,S^m) \geq l}\biggr] \\
    &= \sum_{h \in \mathcal{H}} \sum_{(U,I) \in C_h} \Prob\biggl[ |\hat{\mu}_y(h,U,I,S^m) - \hat{\mu}_h(h,U,I,S^m) - \mu_y(h,U,I) + \mu_h(h,U,I)| > \epsilon \:\bigg|\: {\hat{n}(h,U,I,S^m) \geq l}\biggr] \\
    &\leq \sum_{h \in \mathcal{H}} \sum_{(U,I) \in C_h} \Prob\biggl[| \hat{\mu}_h(h,U,I,S^m) - \mu_h(h,U,I) | + | \mu_y(h,U,I) - \hat{\mu}_y(h,U,I,S^m) | > \epsilon \:\bigg|\: \hat{n}(h,U,I,S^m) \geq l \biggr]
\end{align*}
Since $a + b \geq \epsilon$ implies that either $a \geq \epsilon/2$ or $b \geq \epsilon/2$:
\begin{align*}
    &\sum_{h \in \mathcal{H}} \sum_{(U,I) \in C_h} \Prob\biggl[| \hat{\mu}_h(h,U,I,S^m) - \mu_h(h,U,I) | + | \mu_y(h,U,I) - \hat{\mu}_y(h,U,I,S^m) | > \epsilon \:\bigg|\: \hat{n}(h,U,I,S^m) \geq l \biggr] \\
    &\leq \sum_{h \in \mathcal{H}} \sum_{(U,I) \in C_h} \Prob\biggl[ | \hat{\mu}_h(h,U,I,S^m) - \mu_h(h,U,I) | > \frac{\epsilon}{2} \:\vee\: | \mu_y(h,U,I) - \hat{\mu}_y(h,U,I,S^m) | > \frac{\epsilon}{2} \:\bigg|\: \hat{n}(h,U,I,S^m) \geq l \biggr]
\end{align*}
And by using the union-bound once again:
\begin{align*}
    &\sum_{h \in \mathcal{H}} \sum_{(U,I) \in C_h} \Prob\biggl[ | \hat{\mu}_h(h,U,I,S^m) - \mu_h(h,U,I) | > \frac{\epsilon}{2} \:\vee\: | \mu_y(h,U,I) - \hat{\mu}_y(h,U,I,S^m) | > \frac{\epsilon}{2} \:\bigg|\: \hat{n}(h,U,I,S^m) \geq l \biggr] \\
    &\leq \sum_{h \in \mathcal{H}} \sum_{(U,I) \in C_h} \Prob\biggl[ | \hat{\mu}_h(h,U,I,S^m) - \mu_h(h,U,I) | > \frac{\epsilon}{2} \:\bigg|\:  \hat{n}(h,U,I,S^m) \geq l \biggr] \\
    &\qquad\quad\quad\quad\quad + \Prob\biggl[ | \mu_y(h,U,I) - \hat{\mu}_y(h,U,I,S^m) | > \frac{\epsilon}{2} \:\bigg|\:  \hat{n}(h,U,I,S^m) \geq l  \biggr]
\end{align*}

We would like to use Chernoff inequality (Lemma \ref{abs_chernoff_thm}) to bound the probability with a confidence of $1 - \delta/2$. However,
in order to do so, we must fix the number of samples, $\hat{n}(h,U,I,S^m)$, that $h$ maps to a certain category (rather than using a random variable).
Note that for $\hat{n}(h,U,I,S^m) \geq l $ the probability is maximized at $\hat{n}(h,U,I,S^m) = l$, so we will assume that $\hat{n}(h,U,I,S^m) =l$.  We denote by $S^l|_{(h,U,I)}$ the sub-sample with $[x\in U,h(x)\in I]$, and its size is $l$. 


Now, in order to use Chernoff inequality, we define two random variables, $\hat{Z}_y(h,U,I)$ and $\hat{Z}_h(h,U,I)$, as follows:
\begin{align*}
    &\hat{Z}_y(h,U,I) := \frac{1}{l}\sum_{(x_i, y_i) \in S^l|_{(h,U,I)}}y_i \\
    &\hat{Z}_h(h,U,I) := \frac{1}{l}\sum_{(x_i, y_i) \in S^l|_{(h,U,I)}}h(x_i)
\end{align*}
and we observe that
\begin{align*}
     &\Ex\left[ \hat{Z}_y(h,U,I) \right] = \mu_h(h,U,I)\\
     &\Ex\left[ \hat{Z}_h(h,U,I) \right] = \mu_y(h,U,I)
\end{align*}


Using this notation,
\begin{align*}
    \Prob&\condsb{S^m \in B^m}{S^m \in G^{m,l}} \\
    &\leq \sum_{h \in \mathcal{H}}\sum_{(U,I) \in C_h}\biggl[ 
    \Prob\left[ \left| \hat{Z}_y(h,U,I) - \mu_h(h,U,I) \right| > \frac{\epsilon}{2} \right] + \Prob\left[ \left| \hat{Z}_h(h,U,I) - \mu_y(h,U,I) \right| > \frac{\epsilon}{2} \right]
    \biggr]
    \\
    &\leq \sum_{h \in \mathcal{H}} \sum_{(U,I) \in C_h}{ 4 e^{ - \frac{\epsilon^2}{2} l } }
    \leq \frac{4|\Gamma||\mathcal{H}|}{\lambda} e^{ - \frac{\epsilon^2}{2} l }
\end{align*}

We would like to set $l$ so that $\condprob{S^m \in B^m}{S^m \in G^{m,l}}$ will be at most $\delta/2$, as follows,
\begin{align*}
    &\frac{4|\Gamma||\mathcal{H}|}{\lambda} e^{ - \frac{\epsilon^2}{2} l } \leq \frac{\delta}{2}
    \iff 
    l \geq \frac{ 2 }{ \epsilon^2 } \log \left( \frac{ 8|\Gamma||\mathcal{H}| }{ \delta \lambda } \right)
    \\
\end{align*}
Hence, we set
\[
l = \frac{ 2 }{ \epsilon^2 } \log \left( \frac{ 8|\Gamma||\mathcal{H}| }{ \delta \lambda } \right)
\]

Next, we will bound $\Prob\left[S^m \in \overline{G^{m,l}}\right]$ by $\delta/2$.

Since $m \geq m_{\mathcal{H}}(\epsilon, \delta, \psi, \gamma, \lambda)$ and since $p(h,U,I) \geq \gamma \psi$ for any $h \in \mathcal{H}$ and $(U,I) \in C_h$, we know that for any $h \in \mathcal{H}$ and $(U,I) \in C_h$:
\[
m \geq
\frac{4l}{\gamma \psi}
=
\frac{ 8 \log \left( \frac{ 8|\Gamma||\mathcal{H}| }{ \delta \lambda } \right) }{\epsilon^2 \gamma \psi}
\]

Thus, the expected number of samples we have in each interesting category, is at least twice the value of $l$, i.e.,
\[
\Ex[\hat{n}(h,U,I,S)]
=
m p(h,U,I)
\geq
m \gamma \psi
\geq
2 l
\]

Thus, using the relative version of Chernoff bound, the upper bound we have on $l$, and the lower bound we have on $m$, for any $h \in \mathcal{H}$ and for any interesting category $(U,I) \in C_h$, the probability that $S^m$ has less than $l$ samples in the category $(U,I)$ is bounded by:
\begin{align*}
    &\Prob[\hat{n}(h,U,I,S) \leq l] \leq \Prob\biggl[\hat{n}(h,U,I,S) \leq \frac{\Ex[\hat{n}(h,U,I,S)]}{2}\biggr] \leq e^{-\frac{\Ex[\hat{n}(h,U,I,S)]}{8}} \leq \frac{\lambda \delta}{2|\Gamma||\mathcal{H}|}
\end{align*}

And, by using the union bound:
\begin{align*}
    &\Prob[S^m \in \overline{G^{m,l}}] = \Prob\left[ \exists h \in \mathcal{H}, \exists (U,I) \in C_h : \hat{n}(h,U,I,S) < l \right] \leq |C_h|\frac{\lambda \delta}{2|\Gamma|} \leq \frac{\delta}{2}
\end{align*}

Thus, overall:
\begin{align*}
    &\Prob[S^m \in B^m] \leq \condprob{S^m \in B^m}{S^m \in G^{m,l}} + \Prob[S^m \in \overline{G^{m,l}}] \leq \delta/2 + \delta/2 = \delta
\end{align*}
as required.

\end{proof}

\section{Proofs for Section \ref{sec:Graph}}
\begin{proof}(Proof of Lemma \ref{lma:graph})
\\
Let us assume that $VCdim(\mathcal{H}_v) > d$ and let $S$ be a sample of size $d+1$ such that $\mathcal{H}_v$ shatters $S$.

Let us define the function $f:S \rightarrow \mathcal{Y}$ as:
\[
\forall x \in S: f(x) = v
\]

Let $T \subseteq S$ be an arbitrary subset of $S$.
By assuming that $\mathcal{H}_v$ shatters $S$ we know that there exists $h_v \in \mathcal{H}_v$ such that:
\[
\forall x \in S: h_v(x)=1 \iff x \in T
\]
This means that for the corresponding predictor $h \in \mathcal{H}$:
\[
\forall x \in S: h(x) = v = f(x) \iff x \in T
\]

Thus, using our definition of $f$,
\[
\forall T \subseteq S, \exists h \in \mathcal{H}, \forall x \in S: h(x) = f(x) \iff x \in T
\]
Which means that $S$ is G-shattered by $\mathcal{H}$.
However, since $|S| > d$, it is a contradiction to the assumption that $d_G(\mathcal{H}) \leq d$.
\end{proof}

\begin{proof}(Proof of Lemma \ref{lma:phi})
\\
Assume that $VCdim(\Phi_{\mathcal{H}_v}) > d$ and let $S$ be a sample of $d+1$ domain points and outcomes shattered by $\Phi_{\mathcal{H}_v}$.

Note that $y=0$ implies that $\forall h_v \in \mathcal{H}_v, \forall x \in \mathcal{X}: \phi_{h_v}(x,y) = 0$. Thus, $\forall (x,y) \in S: y=1$ (otherwise $S$ cannot be shattered).

Let $S_x = \{x_j : (x_j, y_j) \in S\}$. Observe that when $y=1$, $\forall h_v \in \mathcal{H}_v, \forall x \in \mathcal{X}: \phi_{h_v}(x,1) = h_v(x)$. Thus, the fact that $S$ is shattered by $\Phi_{\mathcal{H}_v}$ implies that $S_x$ is shattered by $\mathcal{H}_v$.
However, $|S_x| = d+1$. Thus, we have a contradiction to the assumption that $VCdim(\Phi_{\mathcal{H}_v}) > d$.
\end{proof}








\begin{proof}(Proof of Lemma \ref{lma:numDenum})
\\
Let $\mathcal{H}_v$ and $\Phi_{\mathcal{H}_v}$ be the binary prediction and binary prediction-outcome classes of $\mathcal{H}$.

Using Lemmas \ref{lma:graph} and \ref{lma:phi}, and since $d_G(\mathcal{H}) \leq d$, we know that $VCdim(\Phi_{\mathcal{H}_v}) \leq VCdim(\mathcal{H}_v) \leq d$.

In addition, note that:
\begin{align*}
    &\left| \frac{1}{m} \sum_{i=1}^{m}{\indicator{h(x_i) = v}} - \Prob_{x \sim D_U}[h(x) = v] \right| = \left| \frac{1}{m} \sum_{i=1}^{m}{h_v(x_i)} - \Prob_{x \sim D_U}[h_v(x) = 1] \right|,
\end{align*}
And
\begin{align*}
    \biggl| \frac{1}{m} \sum_{i=1}^{m}{\indicator{h(x_i) = v, y=1}} - \Prob_{(x,y) \sim D_U}[h(x) = v, y=1] \biggr| 
    &= \left| \frac{1}{m}  \sum_{i=1}^{m}{\phi_{h,v}(x_i,y_1)} - \Prob_{(x,y) \sim D_U}[\phi_{h,v}(x,y)] \right|.
\end{align*}
and the lemma follows directly from Corollary \ref{fund_cor}.
\end{proof}

\begin{proof}(Proof of Lemma \ref{lma:err})
\\
Let us denote $\xi := \psi\epsilon/3$

\begin{align*}
    \frac{p_1}{p_2} - \frac{\tilde{p}_1}{\tilde{p}_2} &\leq \frac{p_1}{p_2} - \frac{p_1 - \xi}{p_2 + \xi} 
    = \frac{p_1(1 + \xi/p_2)}{p_2(1 + \xi/p_2)} - \frac{p_1 - \xi}{p_2(1 + \xi/p_2)} 
    = \frac{\xi}{p_2(1 + \xi/p_2)} \left[ \frac{p_1}{p_2} + 1 \right]
\end{align*}

Since $p_1, \psi \leq p_2$,
\begin{align*}
    &\frac{\xi}{p_2(1 + \xi/p_2)} \left[ \frac{p_1}{p_2} + 1 \right]
    \leq \frac{\xi}{p_2} \left[ \frac{p_2}{\psi} + \frac{p_2}{\psi} \right]
    = \frac{2\xi}{\psi}
    \leq \frac{3\xi}{\psi}
    = \epsilon.
\end{align*}

Similarly,
\begin{align*}
    \frac{\tilde{p}_1}{\tilde{p}_2} - \frac{p_1}{p_2} &\leq \frac{p_1 + \xi}{p_2 - \xi} - \frac{p_1}{p_2} = \frac{p_1 + \xi}{p_2(1 - \xi/p_2)} - \frac{p_1(1 - \xi/p_2)}{p_2(1 - \xi/p_2)} = \frac{\xi}{p_2(1 - \xi/p_2)} \left[ 1 + \frac{p_1}{p_2} \right].
\end{align*}

Since $p_1, \psi \leq p_2$,
\begin{align*}
    &\frac{\xi}{p_2(1 - \xi/p_2)} \left[ 1 + \frac{p_1}{p_2} \right] 
    \leq \frac{\xi}{p_2(1 - \xi/\psi)} \left[ \frac{p_2}{\psi} + \frac{p_2}{\psi} \right] = \frac{2\xi}{\psi(1 - \xi/\psi)} = \frac{2\epsilon}{3(1 - \epsilon/3)} \leq \frac{2\epsilon}{3(1 - 1/3)} = \epsilon
\end{align*}

Thus,
\[
\left| \frac{p_1}{p_2} - \frac{\tilde{p}_1}{\tilde{p}_2} \right| \leq \epsilon
\]
\end{proof}

\begin{proof}(Proof of Lemma \ref{lma:u_size})
Let $\text{P}_U$ denote the probability of subpopulation $U$:
\[
\text{P}_U := \Prob_{x \sim D}\left[x \in U\right]
\]

Using the relative Chernoff bound (Lemma \ref{rel_chernoff_thm}) and since $\Ex[|S \cap U|] = m \text{P}_U$, we can bound the probability of having a small sample size in $U$. Namely, if $\text{P}_U \geq \gamma$, then:
\begin{align*}
    \Prob_D\left[|S \cap U| \leq \frac{\gamma m }{2}\right] &\leq \Prob_D\left[|S \cap U| \leq \frac{m \text{P}_U }{2}\right]   \leq e^{-\frac{m \text{P}_U}{8}} \leq e^{-\frac{\gamma m}{8}}
\end{align*}

Thus, for any $U \in \Gamma_\gamma$, if $m \geq \frac{8 \log\left(\frac{|\Gamma|}{\delta}\right)}{\gamma}$, 
then, with probability of at least $1 - \frac{\delta}{|\Gamma|}$,
\[
|S \cap U| > \frac{\gamma m}{2}
\]

Finally, using the union bound, with probability at least $1 - \delta$, for all $U \in \Gamma_\gamma$,
\[
|S \cap U| > \frac{\gamma m}{2}
\]
\end{proof}

\begin{proof}(Proof of Theorem \ref{main:thm:graph_ub})

Let $S = \{(x_1, y_1), ..., (x_m, y_m)\}$ be a sample of $m$ labeled examples drawn i.i.d. according to $D$, and let $S_U := \{(x,y) \in S : x \in U\}$ be the samples in $S$ that belong to subpopulation $U$.

Let $\Gamma_\gamma$ denote the set of all subpopulations $U \in \Gamma$ that has probability of at least $\gamma$:
\[
\Gamma_\gamma := \{U \in \Gamma \:|\: \Prob_{x \sim D}[x \in U] \geq \gamma\}
\]

%



Let us assume the following lower bound on the sample size:
\[
m \geq \frac{8 \log\left( \frac{2 |\Gamma|}{\delta} \right)}{\gamma}
\]

Thus, using Lemma \ref{lma:u_size},
we can bound the probability of having a subpopulation $U \in \Gamma_\gamma$ with small number of samples.
Namely, we know that with probability of at least $1 - \delta/2$, for every $U \in \Gamma_\gamma$:
\[
|S_U| \geq \frac{\gamma m}{2}
\]

Next, we would like to show that having a large sample size in $U$ implies accurate approximation of the calibration error, with high probability, for any interesting category in $(U, I)$. For this purpose, let us define $\epsilon', \delta'$ as:
\begin{align*}
    \epsilon' := \frac{\psi \epsilon}{3}
\end{align*}
\begin{align*}
    \delta' := \frac{\delta}{4|\Gamma||\mathcal{Y}|}
\end{align*}

By using Lemma \ref{lma:numDenum} and since $d_G(\mathcal{H}) \leq d$, we know that there exists some constant $a > 0$, such that, for any $v \in \mathcal{Y}$ and any $U \in \Gamma_\gamma$, with probability at least $1 - \delta'$, 
a random sample of $m_1$ examples from $U$, where,
\[
m_1
\geq
a \frac{d + \log(1/\delta')}{\epsilon'^2}
=
9a \frac{d + \log(\frac{4|\Gamma||\mathcal{Y}|}{\delta})}{\epsilon^2 \psi^2}
\]
will have,
\begin{align*}
    &\forall h \in \mathcal{H}: \biggl| \frac{1}{m_1}\sum_{x' \in S_U}{\indicator{h(x') = v}} - \Prob\condsb{h(x) = v}{x \in U} \biggr| \leq  \epsilon' = \frac{\psi \epsilon}{3}
\end{align*}

By using Lemma \ref{lma:numDenum} and since $d_G(\mathcal{H}) \leq d$, we know that for any $v \in \mathcal{Y}$ and any $U \in \Gamma_\gamma$, with probability at least $1 - \delta'$, 
a random sample of $m_2$ labeled examples from $U \times \{0,1\}$, where,
\[
m_2
\geq
a \frac{d + \log(1/\delta')}{\epsilon'^2}
=
9a \frac{d + \log(\frac{4|\Gamma||\mathcal{Y}|}{\delta})}{\epsilon^2 \psi^2}
\]
will have,
\begin{align*}
    &\forall h \in \mathcal{H}: 
    \biggl| \frac{1}{m_2}\sum_{(x', y') \in S_U}{\indicator{h(x') = v, y' = 1}} - \Prob\condsb{h(x)=v, y=1}{x \in U} \biggr| \leq  \epsilon' = \frac{\psi \epsilon}{3}
\end{align*}

Let us define the constant $a'$ in a manner that sets an upper bound on both $m_1$ and $m_2$:
\[
a' := 18 a
\]
and let $m'$ be that upper bound:
\[
m' 
:= 
a' 
\frac{d + \log\left(\frac{|\Gamma||\mathcal{Y}|}{\delta}\right)}
{\psi^2 \epsilon^2}
\geq
\max(m_1, m_2)
\]

Then, by the union bound, if for all subpopulation $U \in \Gamma_\gamma$, $|S_U| \geq m'$, then, with probability at least $1 - 2|\Gamma||\mathcal{Y}|{}\delta' = 1 - \frac{\delta}{2}$:
\begin{align*}
    &\forall h \in \mathcal{H}, \forall U \in \Gamma_\gamma, \forall v \in \mathcal{Y}: \\
        &\quad \biggl| \frac{1}{|S_U|}\sum_{(x',y') \in S_U}{\indicator{h(x') = v}} - \Prob\condsb{h(x) = v}{x \in U} \biggr| \leq \frac{\psi \epsilon}{3} \\
    &\forall h \in \mathcal{H}, \forall U \in \Gamma_\gamma, \forall v \in \mathcal{Y}: \\
        &\quad \biggl| 
        \frac{1}{|S_U|}\sum_{(x', y') \in S_U}{\indicator{h(x')=v, y'=1}} - \Prob\condsb{h(x) = v, y=1}{x \in U} \biggr| \leq  \frac{\psi \epsilon}{3}
\end{align*}

Let us choose the sample size $m$ as follows:
\[
m
:=
\frac{2m'}{\gamma}
=
2a 
\frac{d + \log\left(\frac{|\Gamma||\mathcal{Y}|}{\delta}\right)}
{\psi^2 \epsilon^2 \gamma}
\]

Recall that with probability at least $1 - \delta/2$, for every $U \in \Gamma_\gamma$: 
\[
|S_U| \geq \frac{\gamma m}{2} = m'
\]

Thus, using the union bound once again, with probability at least $1 - \delta$: 
\begin{align*}
\forall h \in \mathcal{H}, \forall U \in \Gamma_\gamma, \forall v \in \mathcal{Y}: \\
    &\quad  \biggl|  \frac{1}{|S_U|}\sum_{x' \in S_U}{\indicator{h(x') = v}} - \Prob\condsb{h(x) = v}{x \in U} \biggr| \leq \frac{\psi \epsilon}{3}
    \\
\forall h \in \mathcal{H}, \forall U \in \Gamma_\gamma, \forall v \in \mathcal{Y}: \\
    &\quad \biggl| \frac{1}{|S_U|}\sum_{(x',y') \in S_U}{\indicator{h(x')=v, y'=1}} - \Prob\condsb{h(x) = v, y = 1}{x \in U} \biggr| \leq \frac{\psi \epsilon}{3}
\end{align*}

To conclude the theorem, we need show that having $\psi\epsilon/3$ approximation to the terms described above, implies accurate approximation to the calibration error.
For this purpose, let us denote:
\begin{align*}
    &p_1(h,U,v) := \Prob\condsb{h(x) = v, y = 1}{x \in U} 
    \\
    &p_2(h,U,v) := \Prob\condsb{h(x) = v}{x \in U} 
    \\
    &\tilde{p}_1(h,U,v) := \frac{1}{|S_U|}\sum_{(x', y') \in S_U}{\indicator{h(x')=v, y'=1}} 
    \\
    &\tilde{p}_2(h,U,v) := \frac{1}{|S_U|}\sum_{x' \in S_U}{\indicator{h(x') = v}}
\end{align*}

Then, with probability at least $1 - \delta$:
\begin{align*}
    &\forall h \in \mathcal{H}, \forall U \in \Gamma_\gamma, \forall v \in \mathcal{Y}: \biggl| \tilde{p}_2(h,U,v) - p_2(h,U,v) \biggr| \leq \frac{\psi \epsilon}{3}
    \\
    &\forall h \in \mathcal{H}, \forall U \in \Gamma_\gamma, \forall v \in \mathcal{Y}: \biggl| \tilde{p}_1(h,U,v) - p_1(h,U,v) \biggr| \leq \frac{\psi \epsilon}{3}
\end{align*}


Using Lemma \ref{lma:err}, 
for all $h \in \mathcal{H}$, $U \in \Gamma_\gamma$ and $v \in \mathcal{Y}$,
if $p_2(h,U,v) \geq \psi$,
then:
\[
\left| 
\frac{p_1(h,U,v)}{p_2(h,U,v)} - 
\frac{\tilde{p}_1(h,U,v)}{\tilde{p}_2(h,U,v)} 
\right|
\leq 
\epsilon
\]

Thus, since
\begin{align*}
    & c(h,U,\{v\}) = \frac{p_1(h,U,v)}{p_2(h,U,v)} -v \\
    & \hat{c}(h,U,\{v\},S)= \frac{\tilde{p}_1(h,U,v)}{\tilde{p}_2(h,U,v)} - v
\end{align*}
then with probability at least $1 - \delta$:
\begin{align*}
    \forall h \in \mathcal{H}, \forall U \in \Gamma, \forall v \in \mathcal{Y}:\qquad
    &\Prob[x \in U] \geq \gamma, \Prob\condsb{h(x) = v}{x \in U} \geq \psi \Rightarrow
    \left| c(h,U,\{v\}) - \hat{c}(h,U,\{v\},S) \right| \leq \epsilon
\end{align*}

\end{proof}

\section{Proofs for Section \ref{sec:lowerBound}}
\begin{proof}(Proof of Theorem \ref{main:thm:lowerBound})
Let $\X=U\cup\{x^2\}$ where $U=\{x^0,x^1\}$ and $x^0\ne x^1$.
Let $H=\{h\}$, where 
\[ 
h(x)=\begin{cases} 
      \frac{1}{2}+\epsilon & x=x^0\\
      0 & else.
\end{cases}
\]
Let $\Gamma=\{U,\{x^2\}\}$.
Let $D\in\{D_1,D_2\}$ where

\[ 
D_1(x,y)= \begin{cases} 
      (1/2+\epsilon)\psi\gamma & (x,y)=(x^0,1) \\
      (1/2-\epsilon)\psi\gamma & (x,y)=(x^0,0) \\
      (1-\psi)\gamma & (x,y)=(x^1,0) \\
      1-\gamma & (x,y)=(x^2,0)
\end{cases}
\]
and
\[ 
D_2(x,y)= \begin{cases} 
      (1/2+\epsilon)\psi\gamma & (x,y)=(x^0,0) \\
      (1/2-\epsilon)\psi\gamma & (x,y)=(x^0,1) \\
      (1-\psi)\gamma & (x,y)=(x^1,0) \\
      1-\gamma & (x,y)=(x^2,0)
\end{cases}
\]

Now we will show a reduction to coin tossing:\\
Consider two biased coins. The first coin has a probability of $r_1=1/2+\epsilon$ for heads and the second has a probability of $r_2=1/2-\epsilon$ for heads. We know that in order to distinguish between the two with confidence $\geq 1- \delta_1$, we need at least $C\frac{\ln(\frac{1}{\delta_1})}{\epsilon^2}$ samples. 

Since 
\[
\Pr_{(x,y)\sim D}[x\in U]=\Pr_{(x,y)\sim D}[x\ne x^2]=\gamma
\]
the first condition for multicalibration holds. 
Now, we use another property of our ``tailor-maded'' distribution $D$ and single predictor $h$, which is $\{x\in \X: h(x)=\frac{1}{2}+\epsilon\}=\{x\in \X: h(x)=\frac{1}{2}+\epsilon,x\in U\}=\{x_0\}$, to get the second condition:
\[
\Pr_D[h(x)=1/2+\epsilon|x\in U]=\Pr_D[x=x^0|x\in U]
=\frac{\psi\gamma}{\gamma}=\psi,
\]
and that
\[
\Pr_D[y=1|h(x)=\frac{1}{2}+\epsilon,x\in U]=\Pr_D[y=1|x=x^0]
\]
is either  $1/2+\epsilon$ (if $D=D_1$) or $1/2-\epsilon$ (in case $D=D_2$) (recall that $D\in\{D_1,D_2\}$). 

Now, if $H$ has the multicalibration uniform convergence property with a sample $S=(x_i,y_i)_{i=1}^m$ of size $m$, and if 
\[\sum_{i=1}^m \frac{\mathbb{I}[y_i=1, h(x_i)=1/2+\epsilon,x_i\in U]}{\sum_{j=1}^m\mathbb{I}[h(x_i)=1/2+\epsilon,x_i\in U]}=
\sum_{i=1}^m \frac{\mathbb{I}[y_i=1, x_i=x^0]}{\sum_{j=1}^m\mathbb{I}[x_i=x^0]}>\frac{1}{2}
\]
holds, then
\[\Pr[y=1|h(x)=\frac{1}{2}+\epsilon,x\in U]=\frac{1}{2}+\epsilon
\]
holds w.p. $1-\delta_1$ (from the definition of multicalibration uniform convergence).

Let us assume by contradiction that we can get multicalibration uniform convergence with $m=\frac{C}{\epsilon^2 \psi \gamma}-\frac{k}{\psi\gamma}<\frac{C}{\epsilon^2 \psi \gamma}$ for some constant $k=\Omega(1)$.

Let $m_{0}$ denote the random variable that represents the number of samples in $S$ such that $x_i=x^0$ (i.e., $h(x_i)=1/2+\epsilon$).
Hence, $\E{m^0}=\gamma\cdot \psi \cdot m= \frac{C}{\epsilon^2}-k$.

From Hoeffding's inequality, $$\Pr[m^0\geq \frac{C}{\epsilon^2}]=\Pr[m^0-\underbrace{(\frac{C}{\epsilon^2}-k)}_{\E{m_0}}\geq
k]\leq e^{-2mk^2} .$$
Let $\delta_2$ be the parameter that holds $e^{-2mk^2}\leq \delta_2$, and let $\delta:=\delta_1+\delta_2$. Then we get that with probability  $> (1-\delta_1)(1-\delta_2)>1-\delta_1-\delta_2=1-\delta$ we can distinguish between the two coins with less than $\frac{C}{\epsilon^2}$ samples, which is a contradiction.


\end{proof}
\end{document}